%% file: ms.tex
\documentclass[10pt,journal,compsoc]{IEEEtran}

\ifCLASSOPTIONcompsoc

  \usepackage[nocompress]{cite}
\else
  \usepackage{cite}
\fi

\ifCLASSINFOpdf

\else

\fi

\usepackage{hyperref}
\usepackage{epsfig}
\usepackage{graphicx}
\usepackage{amsmath}
\usepackage{amssymb}
\usepackage{booktabs,threeparttable}
\usepackage{float}
\usepackage{verbatim}
\usepackage{xcolor}
\usepackage[abs]{overpic}
\newcommand{\igan}{s4GAN}
\newcommand{\mt}{MLMT}
\newcommand\mypara[1]{\vspace{1mm}\noindent\textbf{#1}}
\usepackage{rotating}
\newcommand{\etal}{\textit{et al.}}
\newcommand{\ie}{\textit{i}.\textit{e}.}

\begin{document}

\title{Semi-Supervised Semantic Segmentation\\ with High- and Low-level Consistency}

\author{Sudhanshu Mittal,
        Maxim Tatarchenko,
        and Thomas Brox
\IEEEcompsocitemizethanks{\IEEEcompsocthanksitem The authors are with the Computer Science Department at the University
of Freiburg, Freiburg im Breisgau, Germany.\protect \\

E-mail: \{mittal, tatarchm, brox\}@cs.uni-freiburg.de}}

\IEEEtitleabstractindextext{%
\begin{abstract}
The ability to understand visual information from limited labeled data is an important aspect of machine learning. While image-level classification has been extensively studied in a semi-supervised setting, dense pixel-level classification with limited data has only drawn attention recently. In this work, we propose an approach for semi-supervised semantic segmentation that learns from limited pixel-wise annotated samples while exploiting additional annotation-free images. It uses two network branches that link semi-supervised classification with semi-supervised segmentation including self-training. The dual-branch approach reduces both the low-level and the high-level artifacts typical when training with few labels. The approach attains significant improvement over existing methods, especially when trained with very few labeled samples. On several standard benchmarks - PASCAL VOC 2012, PASCAL-Context, and Cityscapes - the approach achieves new state-of-the-art in semi-supervised learning. 
\end{abstract}

}

% make the title area
\maketitle
\global\csname @topnum\endcsname 0
\global\csname @botnum\endcsname 0

\IEEEdisplaynontitleabstractindextext

\IEEEpeerreviewmaketitle

\ifCLASSOPTIONcompsoc

\input{1_introduction.tex}
\input{2_related_work.tex}

\input{3_method.tex}
\input{4_experiments.tex}
\input{5_conclusions.tex}

\section*{Acknowledgements} This study was supported by the German Federal Ministry of Education and Research via the project Deep-PTL and by the Intel Network of Intelligent Systems. We also thank Facebook for their P100 server donation and gift funding.

\input{ms.bbl}
%{\small
%\bibliographystyle{ieee}
%\bibliography{egbib}
%}

\ifCLASSOPTIONcaptionsoff
  \newpage
\fi

\vfill\eject

\end{document}

%% file: 1_introduction.tex
\section{Introduction}

% trim={<left> <lower> <right> <upper>}

\begin{figure}
\centering
\begin{tabular}{@{}c@{\hspace{1mm}}c@{\hspace{1mm}}c@{\hspace{1mm}}c@{}}
% grid,scale=1,unit=0.4mm,
\begin{overpic}[trim={0 0 0 1.3cm}, clip, width=0.5\linewidth]{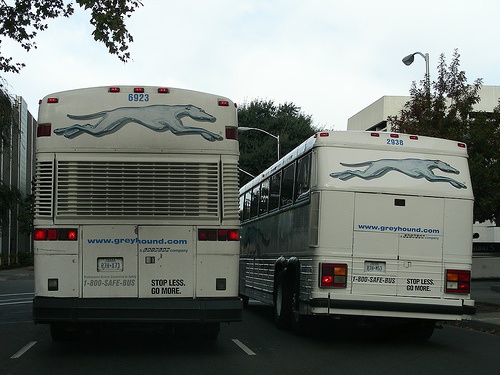}
\put (55.2,2.0) {\color{white}\footnotesize \textbf{(a)}}
\end{overpic} &

\begin{overpic}[trim={0 0 0 1.3cm}, clip, width=0.5\linewidth]{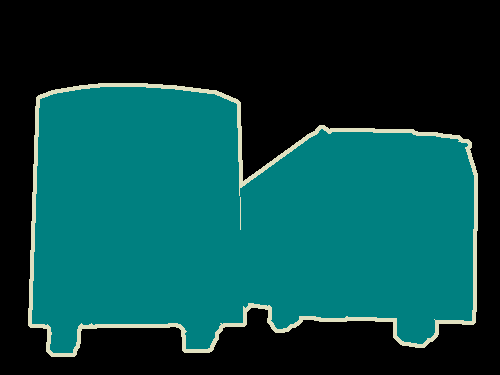}
\put (55.2,2.0) {\color{white}\footnotesize \textbf{(b)}}
\end{overpic} \\

\begin{overpic}[trim={0 0 0 1.3cm}, clip, width=0.5\linewidth]{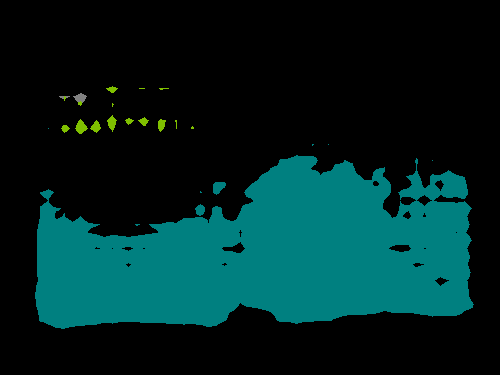}
\put (55.2,2.0) {\color{white}\footnotesize \textbf{(c)}}
\end{overpic} &

\begin{overpic}[trim={0 0 0 1.3cm}, clip, width=0.5\linewidth]{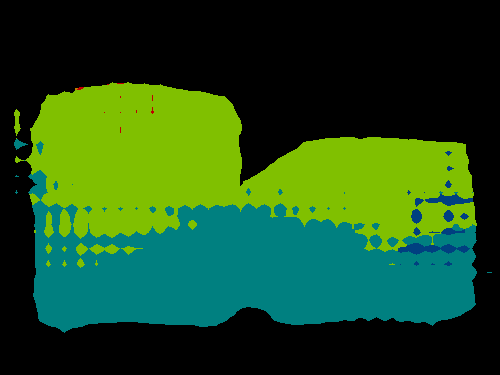}
\put (55.2,2.0) {\color{white}\footnotesize \textbf{(d)}}
\end{overpic} \\

\begin{overpic}[trim={0 0 0 1.3cm}, clip, width=0.5\linewidth]{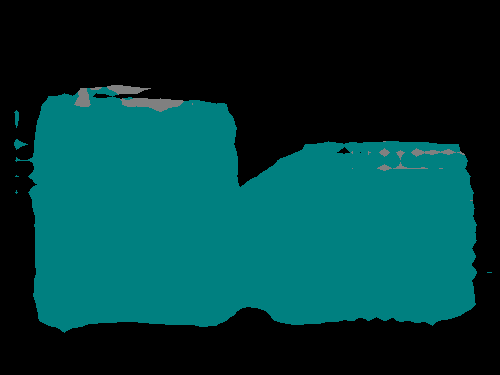}
\put (55.2,2.0) {\color{white}\footnotesize \textbf{(e)}}
\end{overpic} &

\begin{overpic}[trim={0 0 0 1.3cm}, clip, width=0.5\linewidth]{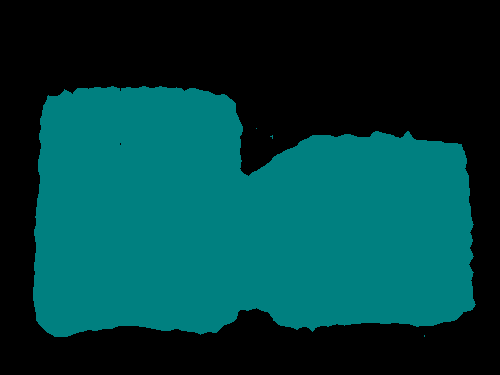}
\put (55.2,2.0) {\color{white}\footnotesize \textbf{(f)}}
\end{overpic} \\

\end{tabular}
%\vspace{1mm}
\caption{An image from the PASCAL VOC dataset (a) and its ground-truth segmentation mask (b). Prediction (c) is obtained with supervised training on 5\% labeled samples. Using the other 95\% unlabeled images, our GAN-based branch improves the shape estimation (d). The second branch adds high-level consistency by removing false positives (e). (f) shows the output when training on 100\% pixel-wise labeled samples.}
% \caption{(a) and (b) show an image and its ground-truth segmentation mask from the PASCAL VOC dataset. Prediction in (c) is obtained with only 5\% labeled samples. Using other 95\% unlabeled images, our GAN-based branch improves the low-level information; see (d) and the second branch removes false positives; see (e). (f) shows the output with 100\% pixel-wise labeled samples.}
%\caption{(a) and (b) show an image and its ground-truth segmentation mask from the PASCAL VOC dataset. Prediction in (c) is obtained with only limited labeled samples. Our GAN-based branch fixes the low-level information (d) and the second branch removes false positives (e). Results are obtained using our approach with 5\% pixel-wise labeled samples with 95\% unlabeled samples. (f) is obtained with 100\% pixel-wise labeled dataset.}
\label{fig:front_page}
\end{figure}

\IEEEPARstart{S}{emantic} segmentation is one of the key computer vision tasks important in various applications including autonomous driving, medical-imaging and robotics. Lately, Deep Convolutional Neural Networks \cite{Lecun98gradient-basedlearning} have demonstrated great results on this task for different datasets \cite{DBLP:journals/pami/ChenPKMY18, Chen_2018_ECCV, Peng_2017_CVPR, Zhang_2018_CVPR}. However, this success usually comes at the cost of collecting dense pixel-wise annotations - a cumbersome process that involves much manual effort.

Attempting to alleviate the problem, several methods exploit weaker forms of supervision: image-level labels \cite{Ahn_2018_CVPR, Pinheiro_2015_CVPR, Wei_2017_CVPR}, bounding boxes \cite{Dai_2015_ICCV, Papandreou_2015_ICCV}, or scribbles \cite{Lin_2016_CVPR, Meng_2018_ECCV}. Only two previous works \cite{Hung_2018_BMVC, Souly_2017_ICCV} have considered true semi-supervised learning for semantic segmentation, which requires having a small subset of fully-labeled samples along with a larger set of completely annotation-free images.

In this work, we propose a dual-branch method for semi-supervised semantic segmentation which can effectively learn from annotation-free samples given a very small set of fully-annotated samples.
Our design is based on the observation that CNNs trained on limited data are subject to two typical modes of failure; see Figure \ref{fig:front_page}(c-d).
The first one appears as inaccuracy in low-level details, such as wrong object shapes, inaccurate boundaries, and incoherent surfaces.
The second one is the misinterpretation of high-level information, which leads to assigning large image regions to wrong classes.

The two network branches are designed to separately address those two types of artifacts. To deal with low-level errors, we propose an improved GAN-based model, where the segmentation network acts as a generator. It is trained together with a discriminator that classifies between generated and ground-truth segmentation maps. Instead of using the original GAN loss, we propose to use the feature matching loss introduced by Salimans \etal \cite{improvedGAN}.
% Moreover, we successfully use self-training within the GAN training dynamics to enhance the segmentation performance, where the GAN's discriminator provides the selection metric for self-training.
Moreover, we introduce the self-training procedure based on the discriminator score which improves the final performance via leveraging high-quality generator predictions as fully labeled samples.
%This strategy allows efficient online bootstrapping using confident predictions. % This strategy allows us to implement an efficient online bootstrapping using good predictions at each iteration. 

For the second type of artifacts, we propose a semi-supervised multi-label classification branch which decides on the classes present in the image and thus aids the segmentation network to make globally consistent decisions. To utilize extra image-level information from unlabeled images, we leverage the success of ensemble-based semi-supervised classification (SSL) methods \cite{Laine2017, NIPS2017_6719}.
% Both branches contribute to the performance improvement and work together in a complementary manner. \tcs {Our method can successfully fix both types of errors while trained with as few as 2\% labeled samples. Figure \ref{fig:front_page} shows a typical example for 5\% labeled data.}

\begin{figure}[H]
\centering
\begin{tabular}{@{}c@{\hspace{1mm}}c@{\hspace{1mm}}c@{\hspace{1mm}}c@{}}
\includegraphics[width=0.5\linewidth]{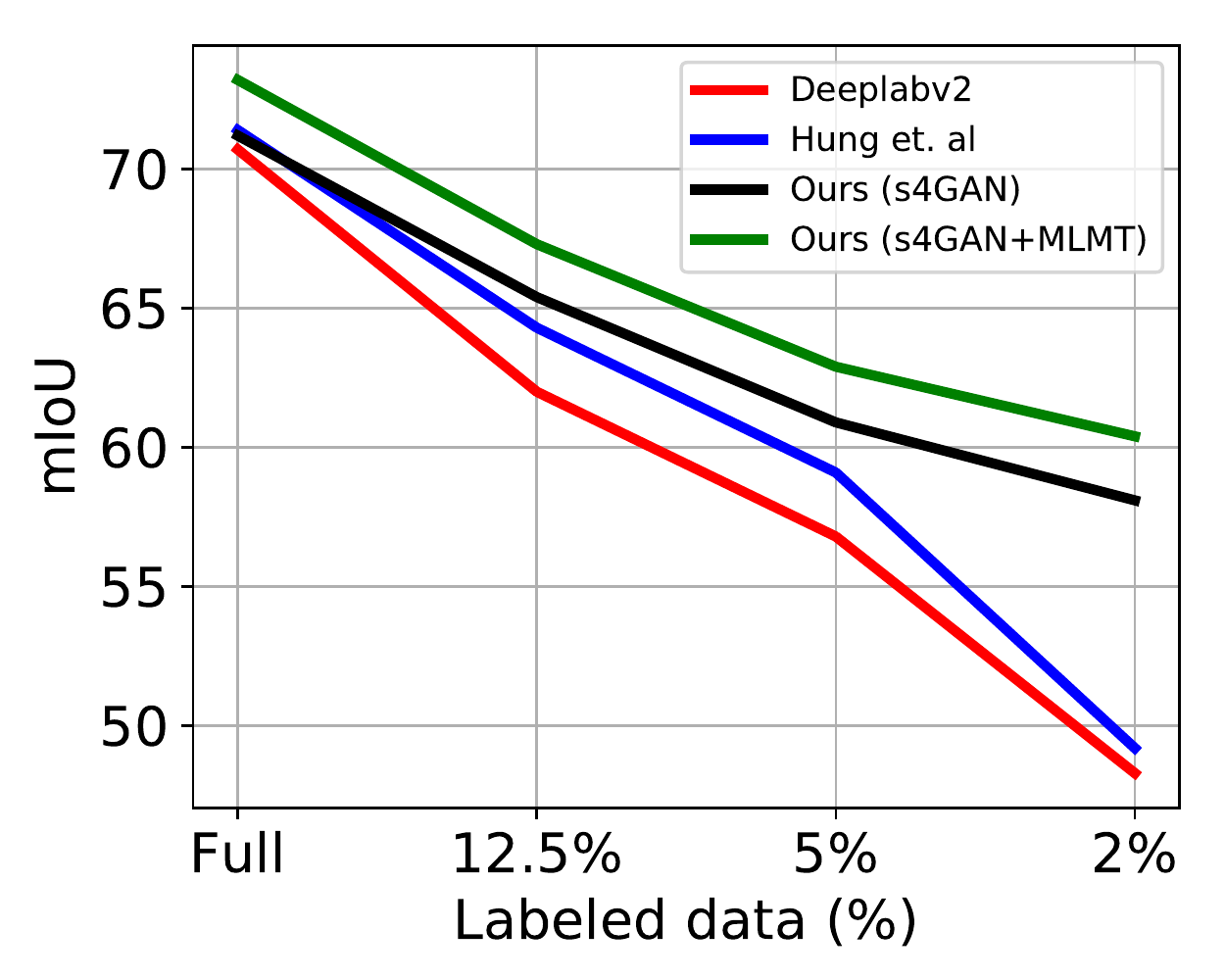} &
\includegraphics[width=0.5\linewidth]{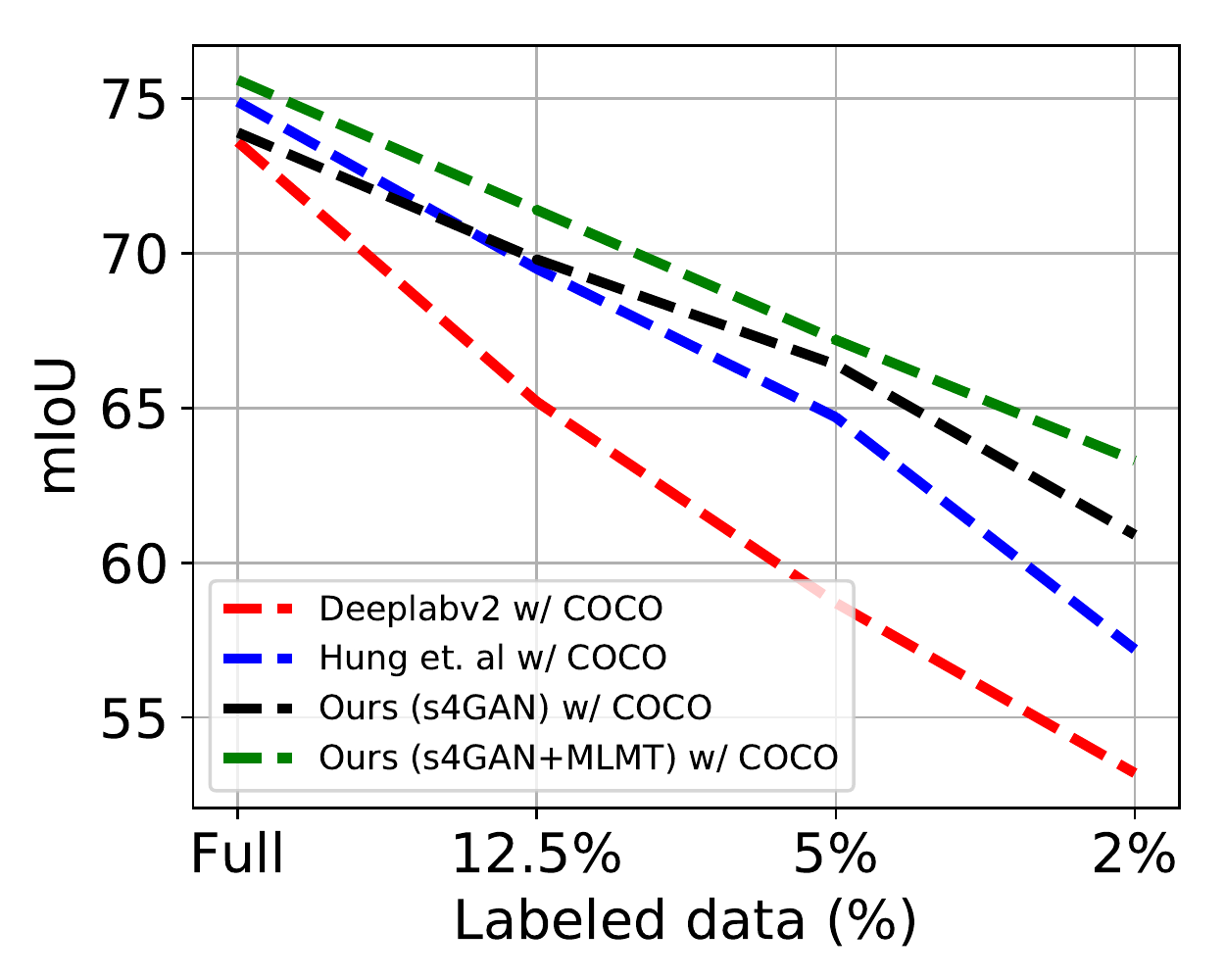} \\
\small (a) & \small (b) 
\end{tabular}
%\vspace{1mm}
\caption{\textbf{Semi-supervised Semantic Segmentation}: The proposed semi-supervised learning (SSL) approach improves over the baselines even when only little labeled data is available using unlabeled data, shows considerable improvement especially with less than 5\% labeled samples. Performance is shown on the PASCAL VOC dataset without (a) and with (b) COCO pre-training.}
\label{fig:results_teaser}
\end{figure}

The two branches act in a complementary manner and successfully fix both low-level and high-level errors; see Fig.~\ref{fig:front_page} for a typical example.
We demonstrate the effectiveness of our approach on different amounts of labeled data across a range of popular semantic segmentation datasets: PASCAL VOC 2012 \cite{pascal_voc}, PASCAL-Context \cite{Mottaghi_2014_CVPR} and Cityscapes \cite{Cordts_2016_CVPR}.
We consistently achieve the best results compared to existing methods and define the new state of the art in semi-supervised semantic segmentation.
Our approach proves particularly efficient when only very few training samples are available: with as little as 2\% labeled data we report an 11\% performance improvement over the state of the art (see Figure \ref{fig:results_teaser}).

% We demonstrate the effectiveness of this approach on different amounts of labeled data across a range of popular semantic segmentation datasets: PASCAL VOC 2012 \cite{pascal_voc}, PASCAL-Context \cite{Mottaghi_2014_CVPR}, and Cityscapes \cite{Cordts_2016_CVPR}. Oliver \etal \cite{oliver2018realistic} show that when labeled data is scarcely available, transfer learning via pre-training on suitable labeled dataset may be a good alternative to achieve high performance. However, for certain applications like bio-medical imaging, such suitable pre-training datasets are rarely available. Therefore, we show that our approach yields clear benefits both in the case where the base segmentation network was pre-trained on a large dataset like COCO, \emph{and} in the case where the network was not pre-trained on such dataset. 

%\TODO{Maybe say something about the realistic evaluation}

%The approach yields clear benefits both in the case where the base segmentation network was pre-trained on a large dataset like COCO, \emph{and} in the case where the network was not pre-trained on such dataset.  

% We further show that the approach can exploit extra image-level weak annotations when those are available, which leads to superior performance compared to existing methods.
We further show that the approach can easily make use of extra image-level weak annotations when those are available. 
It compares favorably to the existing methods operating in the same setting. The source code of this paper is available \footnote{Source code: \url{https://github.com/sud0301/semisup-semseg}}.

%
%Here, we briefly summarize the contributions of this work.
%\begin{itemize}
%    \item We propose a dual-branch semi-supervised semantic segmentation approach which achieves state-of-the-art results on all the labeled to unlabeled data splits on three different datasets. 
%
%    \item We propose a new improved GAN-based approach which consistently performs well even with few labeled samples, without using extra weak labels and without pre-training the network using other segmentation datasets. % 
%    
%    \item  We extend semi-supervised classification for semi-supervised multi-label classification to further improve segmentation performance.
%    
%    \item Our method can make use of extra weak image-level annotations to enhance its performance, thus achieve favorable results as compared to other semi-supervised segmentation approaches with extra weak annotations.
    % Competitive or state-of-the-art
    
%\end{itemize}

\begin{comment}
\begin{figure}[t]
    \includegraphics[width=\columnwidth]{./images/p1b.png}
    \caption{Contribution figure}
\end{figure}
\end{comment} 

%% file: 2_related_work.tex
\section{Related Work}
\label{sec:related_work}

\begin{figure*}[t]
    \centering
    %\includegraphics[width=\textwidth]{./dummy_images/pipeline_11.pdf}
    % trim={<left> <lower> <right> <upper>}
    \includegraphics[trim={0 2.6cm 3.5cm 1.2cm}, clip, width=\textwidth]{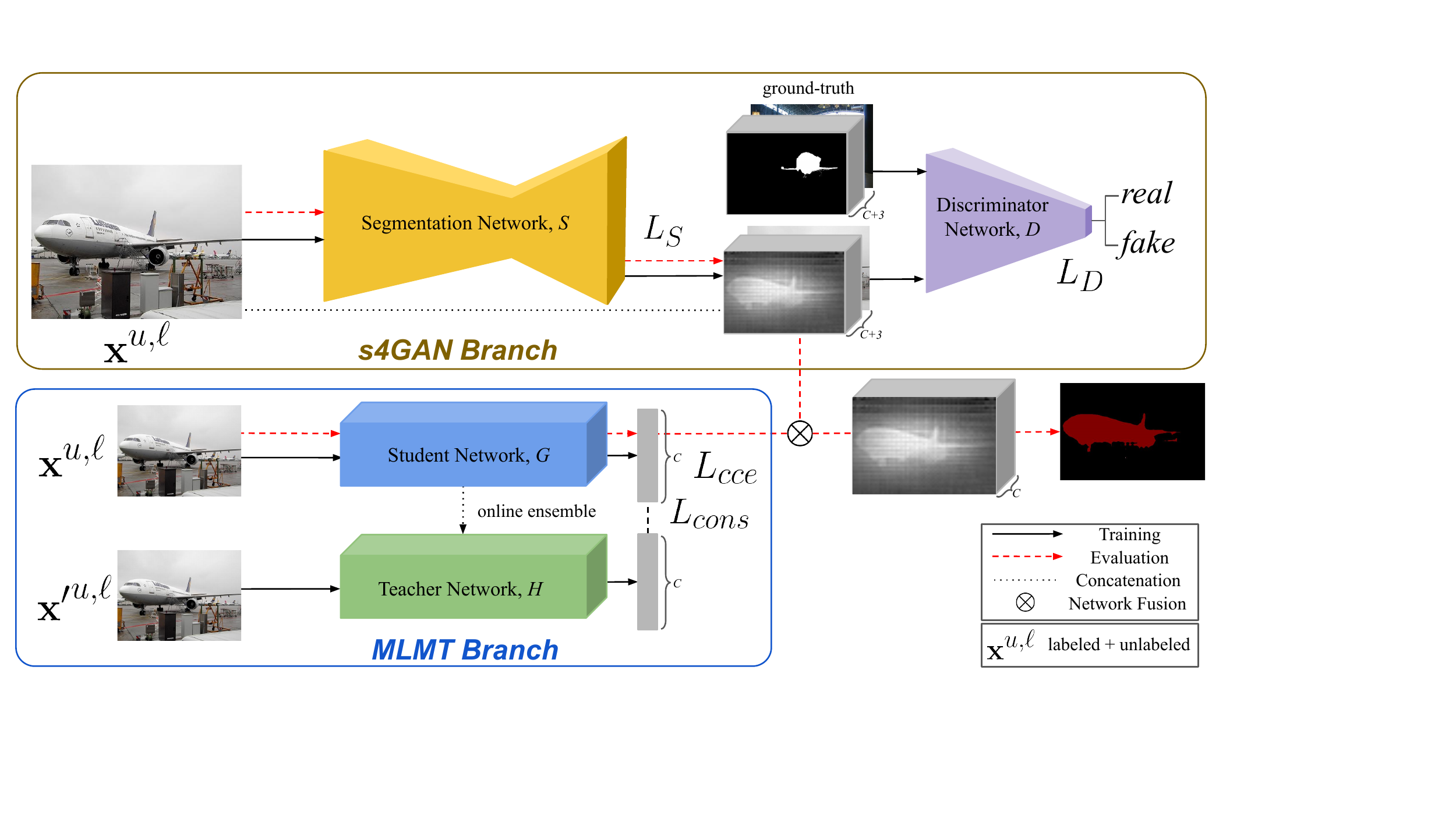}
    \vspace{-5mm}
    \caption{Overview of our proposed semi-supervised segmentation approach. The \igan{} branch is a GAN-based model which improves the low-level details in the segmentation prediction.
    % The \mt{} branch does semi-supervised multi-label classification to improve the class-level information by removing false-positive class predictions.
    The \mt{} branch performs semi-supervised multi-label classification to exploit class-level information for removing false-positive predictions from the segmentation map.
    } 
    \label{fig:pipeline}
\end{figure*}

%Weakly-supervised semantic segmentation approaches often tend to produce smooth segments or make use of computationally expensive post-processing techniques based on probabilistic graphical models like CRFs \cite{NIPS2011_4296}. Semi-supervised approaches aim to resolve the ambiguities by using few labeled samples.

%\textbf{Weakly-supervised and Semi-supervised Segmentation:} \TODO{Double check this paragraph} To reduce annotation effort, most existing approaches rely on weakly-supervised training schemes, i.e., a part of the training samples does not come with a full segmentation but only image-level class labels \cite{Papandreou_2015_ICCV, Wei_2018_CVPR} or bounding boxes \cite{Dai_2015_ICCV, Khoreva_2017_CVPR, Papandreou_2015_ICCV}.

% Sudhanshu Version
\textbf{Weakly-supervised and Semi-supervised Segmentation.} To reduce annotation effort, most existing approaches rely on weakly- and semi-supervised training schemes which use weak labels from the whole dataset like image-level class labels \cite{Papandreou_2015_ICCV, Wei_2018_CVPR}, bounding boxes \cite{Dai_2015_ICCV, Khoreva_2017_CVPR, Papandreou_2015_ICCV} or scribbles \cite{Lin_2016_CVPR, Meng_2018_ECCV}, where semi-supervised schemes \cite{Khoreva_2017_CVPR, Papandreou_2015_ICCV, Wei_2018_CVPR} additionally use a few pixel-wise segmentation labels.

Only two recent works \cite{Hung_2018_BMVC, Souly_2017_ICCV} consider true semi-supervised learning, i.e., they improve semantic segmentation with completely annotation-free images. These methods, like ours, utilize a GAN-based model. However, both approaches use the GAN in a different manner. Souly \etal \cite{Souly_2017_ICCV} use the GAN to generate additional images to enhance the features learned by the segmentation network. They further extend their semi-supervised method by generating additional class-conditional images.

Most related to ours is the work by Hung \etal \cite{Hung_2018_BMVC}.
They also propose a GAN-based design which enables learning from unlabeled samples. However, our framework is substantially different in many details. 
Hung \etal use an FCN-based~\cite{Long_2015_CVPR} discriminator which yields a dense probabilistic map for each pixel, whereas we propose an image-wise discriminator. In contrast to the two-stage training process of \cite{Long_2015_CVPR}, we propose an automatic integration of self-training based on the GAN training dynamics. 
Moreover, we propose to use a feature matching loss, which is crucial for the stability of GAN training, especially when only few labeled samples are available. Finally, we add a semi-supervised multi-label classification branch for resolving high-level inconsistencies.

%whereas our discriminator yields a single confidence value for the whole prediction and the segmentation network is used as generator. Additionally, we make use of different loss functions and training procedure.} 

Also Luc \etal \cite{Luc2016SemanticSU} share some common ground with our work, although their work does not comprise semi-supervised learning. In their case the GAN replaces CRF-post-processing which enhance low-level consistency in the segmentation maps. Luc \etal \cite{Luc2016SemanticSU} optimize the original GAN loss to encourage predicted segmentation maps to be similar to the ground-truth maps and show that it improves the performance in a fully-supervised setting.

%In this work, we propose an improved GAN-based approach which can also learn using very limited amount of fully-annotated samples along with other annotation-free samples without pre-training with any segmentation dataset whereas previous method  \cite{Hung_2018_BMVC} makes use of a pre-training segmentation dataset before applying their semi-supervised approach. 

\textbf{Semi-supervised Classification.} In contrast to segmentation, many semi-supervised methods exist for image classification \cite{athiwaratkun2018improving, Laine2017, Miyato2018VirtualAT, improvedGAN, NIPS2017_6719}. Oliver \etal \cite{oliver2018realistic}, however, criticize that most of the work lacks realistic evaluation to address real-world conditions. They propose a new experimental methodology closer to the real-world settings. We find that consistency-based semi-supervised classification methods \cite{athiwaratkun2018improving, NIPS2017_6719} show improvement over the supervised baseline while satisfying at least two procedures mentioned by Oliver \etal \cite{oliver2018realistic}. Firstly, those methods show improvement over the supervised setting while using a high-quality supervised baseline. Secondly, they can improve upon the pre-trained network using unlabeled data. We use the Mean-Teacher method \cite{NIPS2017_6719} in our approach. 

\textbf{Network Fusion.} The approach to fuse spatial and class information by channel-wise selection is inspired by some recent works in other domains. Hu \etal \cite{hu2018senet} proposed SE-Net for image classification, which learns to combine spatial and channel-wise information by calibrating channel-wise features maps. Following SE-Net, Zhang \etal \cite{Zhang_2018_CVPR} proposed to incorporate class information in semantic segmentation to highlight class-dependent feature maps. Multiple works \cite{NIPS2015_5858, Wei_2017_CVPR, Wei_2018_CVPR} have explored the usage of classification methods, both in a shared and a decoupled manner to constructively use class information for semi- and weakly supervised semantic segmentation. In this work, we use a decoupled approach with late fusion of spatial and class information to remove false positive class channels.

%In this work, we leverage one of the latest approaches and extend it for multi-label semi-supervised classification to discourage false positive predictions made by the segmentation network.

%% file: 3_method.tex
\section{Method}

We propose a two-branch approach to the task of semi-supervised semantic segmentation as shown in Figure \ref{fig:pipeline}. The lower branch predicts pixel-wise class labels and is referred to as the \textit{Semi-Supervised Semantic Segmentation GAN} (\igan{}). The upper branch performs image-level classification and is denoted as the \textit{Multi-Label Mean Teacher} (\mt{}).

% The objective of the IGAN branch is to produce segmentation maps with improved shape, coherent surfaces and refined boundaries. In our GAN model, the generator learns to produce segmentation maps similar to the ground truth masks using the unsupervised feature matching loss and discriminator learns to classify input map into ground-truth mask as 1 or generated segmentation map as 0. The discriminator receives a concatenated input of original images and its corresponding segmentation map, thus it is able to gauge the quality of the prediction. This quality measure is further used for self-training using confident predictions to improve the performance of the prediction.

The core of the \igan{} branch is a standard segmentation network for generating per-pixel class labels given the input image.
We combine conventional supervised training with adversarial training, which allows leveraging unlabeled data to improve the prediction quality.
The segmentation network acts as a generator and is trained together with a discriminator responsible for distinguishing the ground truth segmentation maps from the generated ones.
We additionally treat the outputs of the discriminator as a quality measure and use it to identify the best predictions which are further exploited for self-training.

The \mt{} branch predicts image-level class labels used to filter the \igan{} outputs.
Its core is a Mean Teacher classifier, which effectively removes false positive predictions of the segmentation network.
% The \mt{} branch performs semi-supervised multi-label classification to make the classifier more robust to false-positive predictions as compared to a supervised classifier trained only on a small labeled dataset. The objective of the MLMT branch is to remove the false positive predictions made by the segmentation network.
% For evaluation, the output of the two branches are simply fused by channel-wise selection, explained further in detail in Section \ref{sec:net_fuse}. % The generator of the IGAN branch produces $C$ segmentation maps, one for each class and the output the MLMT branch is a probabilistic $C$ dimensional vector, each dimension indicating the presence of each class. During fusion, the segmentation maps, which are obtained from IGAN, with low corresponding class probability from MLMT branch are removed.
The contributions of the two branches are complementary to each other. Their outputs are combined to produce the final result.

\textbf{Notations:} Our dataset $\mathcal{D}$ is split into the labeled part $\mathcal{D}^\ell=\{\mathbf{x}^\ell, \mathbf{y}^\ell\}$ and the unlabeled part $\mathcal{D}^u=\{\mathbf{x}^u\}$, where $\mathbf{x}$ are the input images and $\mathbf{y}$ are the pixel-wise segmentation labels. 

\subsection{\igan{} for Semantic Segmentation}
In our \igan{} model, the segmentation network $S$ acts as a generator network that takes image $\mathbf{x}$ as input and predicts $C$ segmentation maps, one for each class. The discriminator $D$ gets the concatenated input of the original image and its corresponding predicted segmentation. Its task is to match the distribution statistics of the predicted and the real segmentation maps. 

\subsubsection{Training S}
\label{sec:training_s}
The segmentation network $S$ is trained with loss $L_S$, which is a combination of three losses: the standard cross-entropy loss, the feature matching loss, and the self-training loss.

%%----------------------------------------CE Loss-----------------------------------------
\textbf{Cross-entropy loss.}
This is a standard supervised pixel-wise cross-entropy loss term $L_{ce}$. The loss for the output $S(\mathbf{x})$ of size $H \times W \times C$ is evaluated only for the labeled samples $\mathbf{x}^\ell$:
\begin{equation}
    L_{ce} = - \sum\limits_{\substack{h, w, c}} \mathbf{y}^\ell(h,w,c)\log S(\mathbf{x}^\ell)(h,w,c),
\end{equation}
where $\mathbf{y}^\ell$ is the ground-truth segmentation mask.
%%----------------------------------------FM Loss ----------------------------------------

\textbf{Feature matching loss.}
The feature matching loss $L_{fm}$ \cite{improvedGAN} aims to minimize the mean discrepancy between the feature statistics of the predicted, $S(\mathbf{x}^u)$ and the ground-truth segmentation maps, $\mathbf{y^\ell}$:
% We use the feature matching (FM) loss, as proposed by Salimans \etal \cite{improvedGAN}, to optimize the generator/segmentation network ($S$). Using feature matching loss, the generator minimizes the mean discrepancy between the features of the predicted segmentation maps for unlabeled samples $\mathbf{x}^u$ and the ground-truth segmentation masks $\mathbf{y}^\ell$. These features are defined by the intermediate layer $k$ of the discriminator network $D$. The FM loss term, $L_{fm}$, for the segmentation network is defined as: 
\begin{align}
 L_{fm} =& \rVert \mathbb{E}_{(\mathbf{x}^\ell,\mathbf{y}^\ell)\sim \mathcal{D}^\ell} [D_k(\mathbf{y}^\ell \oplus \mathbf{x}^\ell)] \notag \\ &
  -\mathbb{E}_{\mathbf{x}^u\sim \mathcal{D}^u} [D_k(S(\mathbf{x}^u) \oplus \mathbf{x}^u)] \lVert,
\end{align}
where $D_k(\cdot)$ is the intermediate representation of the discriminator network after the $k^{th}$ layer. Both ground-truth and predicted segmentation masks are concatenated with their corresponding input images. %$p_{gt}$ is the probability distribution of the ground-truth segmentation maps and $p_{data}$ is the input data distribution.
Intuitively, it encourages the generator to predict segmentation maps which have the same feature statistics as the ground truth, and therefore also qualitatively resemble the ground truth. This loss is used on the unlabeled samples $\mathbf{x}^u$, thus forcing plausible solutions even for cases where dense labels are unavailable.
% In the semi-supervised setting, samples from $p_{gt}$ are only available for limited amount of labeled data, whereas samples from $p_{data}$ are available for the whole dataset.\\

%%-----------------------------------------ST Loss-----------------------------------------
\textbf{Self-training loss.} During GAN training, the discriminator ($D$) and the generator ($G$) networks need to be balanced. If $D$ starts off being too strong, it does not provide any useful learning signal for $G$. In order to facilitate such balanced dynamics, we introduce the self-training (ST) loss. The main idea is to pick the best generator outputs (i.e. those able to fool $D$) which do not have the corresponding ground truth, and reuse them for supervised training. Intuitively, this pushes $G$ more to produce predictions which $D$ cannot distinguish from the real ones. This impedes the progress of $D$ and does not allow it to become too strong.

Technically, the output of $D$ varies between 0 and 1, where 0 should be assigned to the predicted segmentation maps and 1 to the ground-truth segmentation maps.
We use this score as a confidence measure for the quality of the predicted segmentations.
High-quality predictions are used for supervised training, i.e. we calculate the standard cross-entropy loss based on them.
The self-training loss term $L_{st}$ is thus defined as: 
\begin{align}
    L_{st} &=
    \begin{cases}
   -\sum\limits_{h,w,c}\mathbf{y}^*\log S(\mathbf{x}^u),   & \text{if } D(S(\mathbf{x}^u) \geq \gamma\\
    0,                              & \text{otherwise},
    \end{cases}
\end{align}
 %& \text{where } \mathbf{y}^*= V * \argmax_{c \in C} S(\mathbf{x}^u),
where $\gamma$ is the confidence threshold which controls how certain $D$ needs to be about the prediction in order for it to be used in self-training; $\mathbf{y}^*$ are the pseudo pixel-wise labels generated from the prediction $S(\mathbf{x}^u)$ of the segmentation network.

The final training objective $L_{S}$ is composed of the three described terms:
\begin{equation}
    L_{S} = L_{ce} + \lambda_{fm} L_{fm} + \lambda_{st}L_{st},
\end{equation}
where $\lambda_{fm}, \lambda_{st} > 0 $ are the corresponding weights.

%%--------------------------------------GAN Discriminator------------------------------------

\subsubsection{Training D}
The discriminator network is trained with the original GAN objective as proposed by Goodfellow \etal \cite{NIPS2014_5423}
\begin{align}
    L_{D} =& \mathbb{E}_{(\mathbf{x}^\ell, \mathbf{y}^\ell)\sim \mathcal{D}^\ell}[\log D(\mathbf{y}^\ell \oplus \mathbf{x}^\ell)] \notag\\&+
    \mathbb{E}_{\mathbf{x}^u\sim \mathcal{D}^u}[\log(1- D(S(\mathbf{x}^u) \oplus \mathbf{x}^u))],
\end{align}
where $\oplus$ denotes concatenation along the channel dimension.
Following the original GAN idea, $D$ learns to differentiate between the real $\mathbf{y}^\ell$ and the fake segmentation masks $S(\mathbf{x}^u)$ concatenated with the corresponding input images.
% The discriminator receives the original image concatenated channel-wise with: one-hot ground-truth segmentation masks, $(\mathbf{y}^\ell \oplus \mathbf{x}^\ell)_{H \times W \times C+3}$ as a real sample and predicted segmentation maps, ($S(\mathbf{x}^u) \oplus \mathbf{x}^u)_{H \times W \times C+3}$ as a fake sample.

% The discriminator gets either a concatenated input of original image and its corresponding ground-truth segmentation mask ($(\mathbf{y}^\ell \oplus \mathbf{x}^\ell)_{H \times W \times C+3}$) or the concatenated input of original image and its corresponding predicted segmentation maps ($S(\mathbf{x}^u) \oplus \mathbf{x}^u)_{H \times W \times C+3}$). 

%%---------------------------------------Semi-supervised Classification-------------------------

\subsection{Multi-label Semi-supervised Classification}

We extend an ensemble-based semi-supervised classification method (Mean Teacher) \cite{NIPS2017_6719} for semi-supervised multi-label image classification.
This model consists of two networks: a student network $G$ and a teacher network $H$.
Both networks receive the same images under different small perturbations.
The weights ($\theta'$) of the teacher network are the exponential moving average (online ensemble) of the student network's weights ($\theta$).
The predictions made by the student model are encouraged to be consistent with the predictions of the teacher model using the consistency loss which is the mean-squared error between the two predictions.

We optimize the student network using the categorical cross-entropy loss $L_{cce}$ for labeled samples $\mathbf{x}^\ell$, and using the consistency loss $L_{cons}$ for all available samples ($\mathbf{x}^{u, \ell}$): 
% \begin{align}
%     L_{MT} &=  \underbrace{-\sum_{c} \mathbf{z}^\ell(c) \log(G(\mathbf{x}^\ell;\theta)(c))}_\text{$L_{cce}$} \notag\\
%     &+ \lambda_{cons}\underbrace{ \lVert G(\mathbf{x}^{(u, \ell)};\theta) - H(\mathbf{x'}^{(u, \ell)};\theta') \rVert^2}_\text{$L_{cons}$},
% \end{align}
\begin{align}
    L_{MT} =&  \underbrace{-\sum_{c} \mathbf{z}^\ell(c) \log(G_\theta(\mathbf{x}^\ell)(c))}_\text{$L_{cce}$} \notag\\
    &+ \lambda_{cons}\underbrace{ \lVert G_\theta(\mathbf{x}^{(u, \ell)}) - H_{\theta'}(\mathbf{x'}^{(u, \ell)}) \rVert^2}_\text{$L_{cons}$},
\end{align}
where $\mathbf{x}$ and $\mathbf{x}'$ are differently augmented images for student and teacher network respectively, $\mathbf{z}^\ell$ is the multi-hot vector for ground-truth class labels. The parameter $\lambda_{cons} >0$ controls the weight of the consistency loss in $L_{MT}$. 

%%---------------------------------------Network Fusion--------------------------------------------

\subsection{Network Fusion}\label{sec:net_fuse}
The two described branches are trained separately.
For evaluation, the output of the classification branch simply deactivates the segmentation maps of those classes not present in the input image:
\begin{align}
    S(\mathbf{x})_c &= 
    \begin{cases}
     0                          & \text{if } G(\mathbf{x}_c) \leq \tau \\
     S(\mathbf{x})_c            & \text{otherwise}
    \end{cases}
\end{align}
where $S(\mathbf{x})_c$ is the segmentation map for class $c$, $G(\mathbf{x})_c$ is the soft output of the MLMT-branch, and $\tau=0.2$ is a threshold on that soft output obtained by cross-validation.

%\begin{align}
%    \mathbf{y_i} &=  \mathbf{y_i^g} \odot \mathbf{y_i^\ell} & \text{if } D(S(\mathbf{x}) \geq \tau\\
%\end{align}

%% file: 4_experiments.tex
\section{Experiments}
The proposed approach was evaluated on the PASCAL VOC 2012 segmentation benchmark, the PASCAL-Context dataset, and the Cityscapes dataset. 
%%------ DATASET and IMPLEMENTATION DETAILS ------------------%%

% \subsection{Dataset and Implementation Details}

\begin{figure*}[h!]
\centering
\begin{tabular}{c@{\hspace{1mm}}c@{\hspace{1mm}}c@{\hspace{1mm}}c@{\hspace{1mm}}c}
(a) Original & (b) Ground Truth & (c) Baseline & (d) Hung \etal \cite{Hung_2018_BMVC} & (e) Our Results  \\[6pt]
 \includegraphics[trim={0 2.5cm 0 0}, clip, width=34mm]{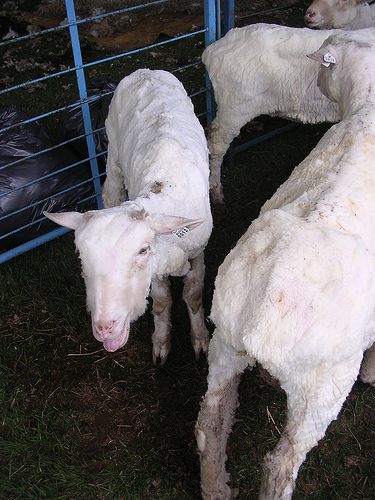} &
 \includegraphics[trim={0 2.5cm 0 0}, clip, width=34mm]{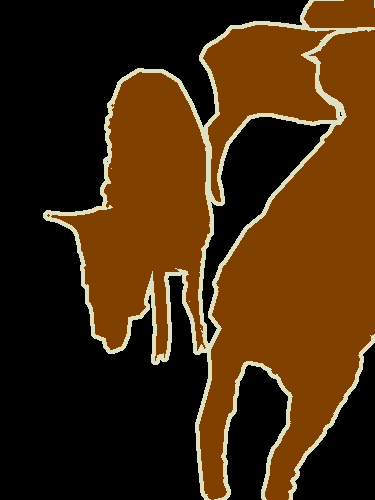} &
 \includegraphics[trim={0 2.5cm 0 0}, clip, width=34mm]{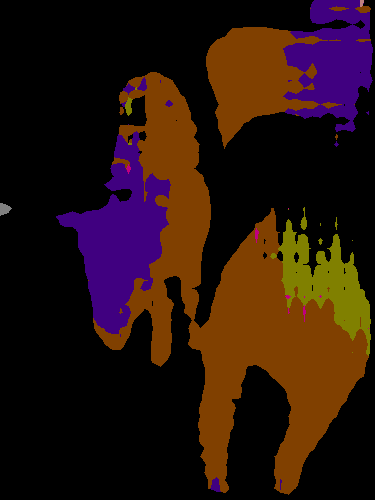} &
 \includegraphics[trim={0 2.5cm 0 0}, clip, width=34mm]{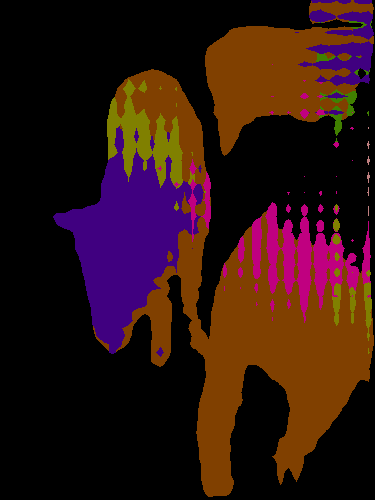} &
 \includegraphics[trim={0 2.5cm 0 0}, clip, width=34mm]{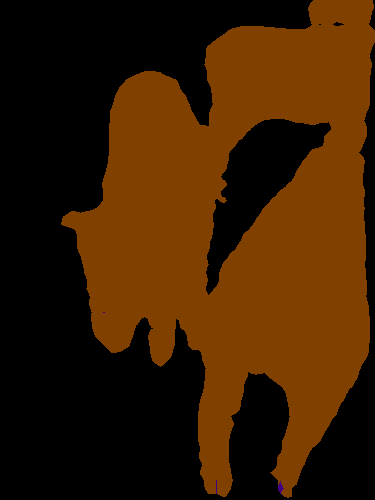}  \\
 \includegraphics[width=34mm]{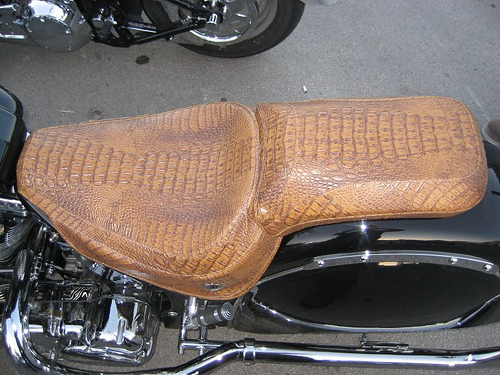} &
 \includegraphics[width=34mm]{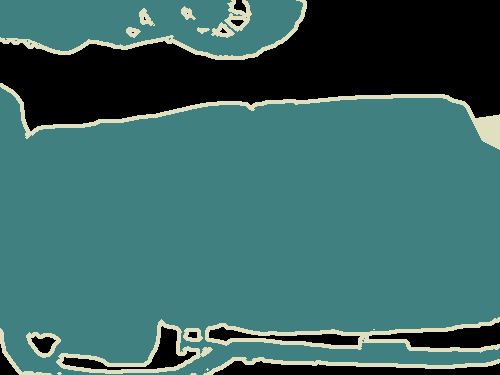} &
 \includegraphics[width=34mm]{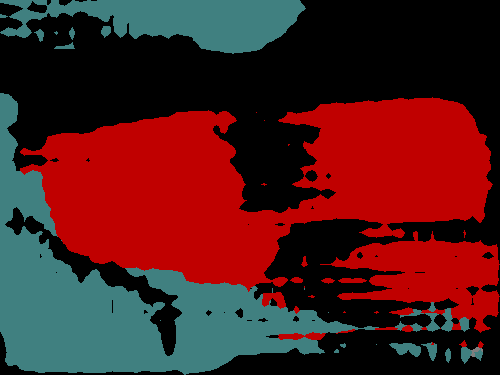} &
 \includegraphics[width=34mm]{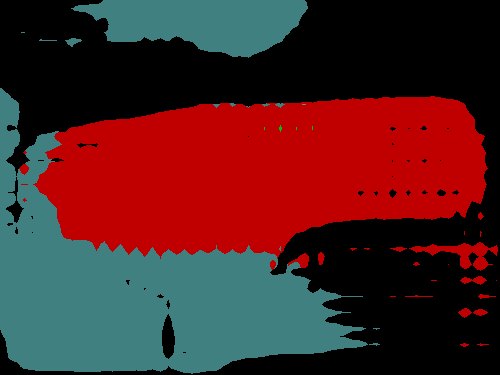} &
 \includegraphics[width=34mm]{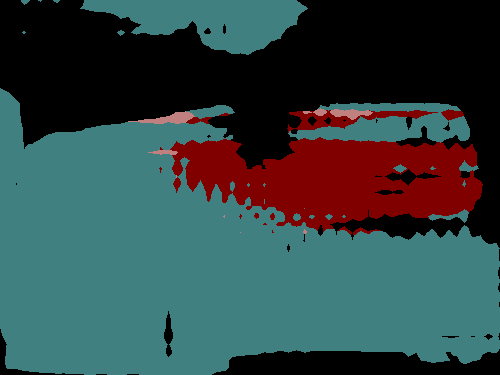}  \\
 
 \includegraphics[trim={0 0.5cm 0 1.5cm}, clip, width=34mm]{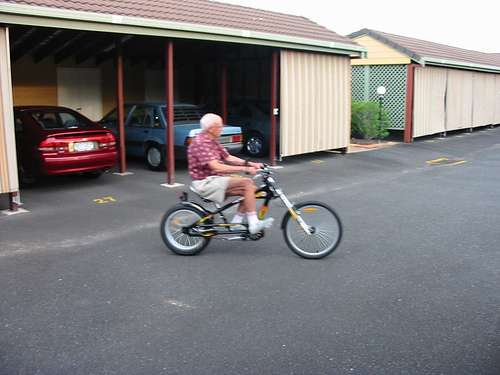} &
 \includegraphics[trim={0 0.5cm 0 1.5cm}, clip, width=34mm]{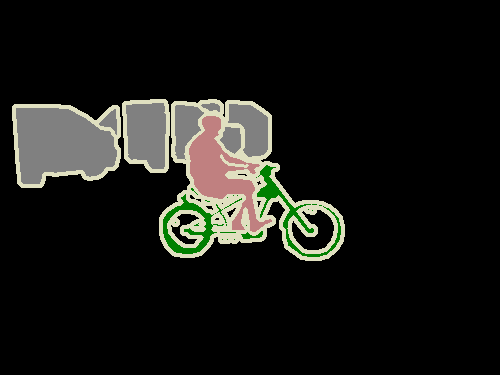} &
 \includegraphics[trim={0 0.5cm 0 1.5cm}, clip, width=34mm]{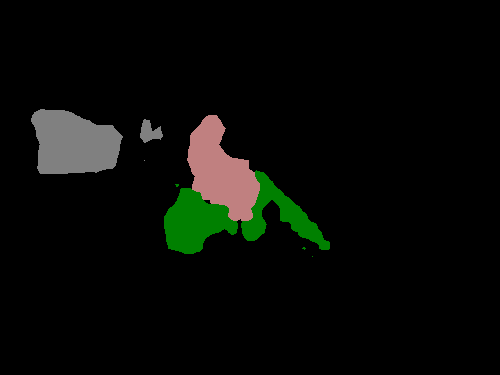} &
 \includegraphics[trim={0 0.5cm 0 1.5cm}, clip, width=34mm]{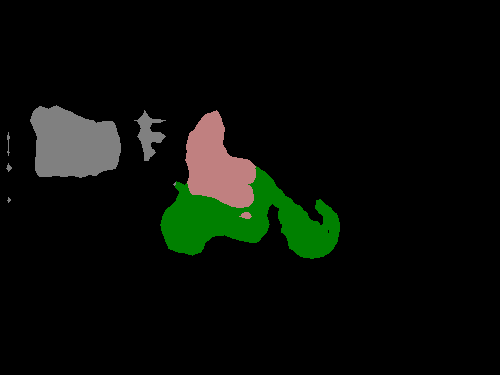} &
 \includegraphics[trim={0 0.5cm 0 1.5cm}, clip, width=34mm]{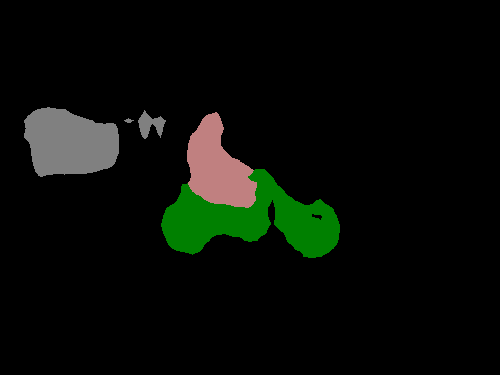}  \\
 
 \includegraphics[trim={0 0.5cm 0 0}, clip, width=34mm]{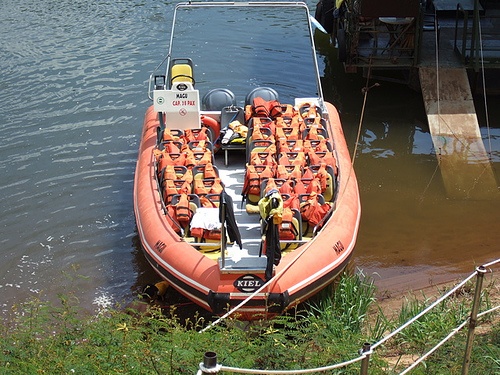} &
 \includegraphics[trim={0 0.5cm 0 0}, clip, width=34mm]{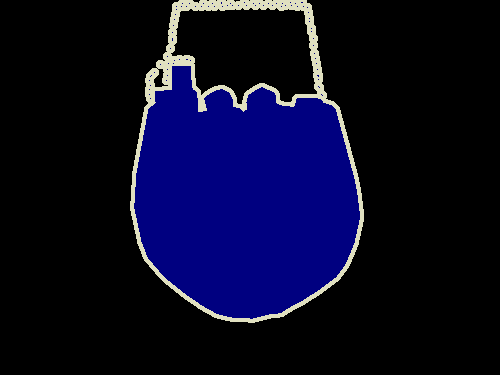} &
 \includegraphics[trim={0 0.5cm 0 0}, clip, width=34mm]{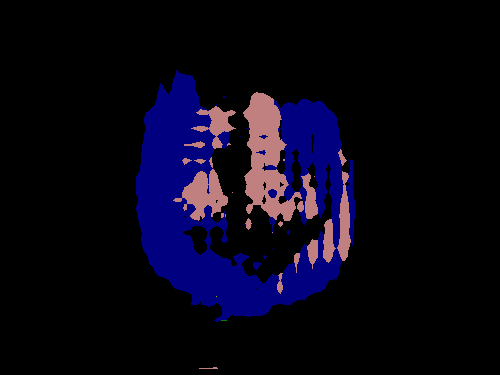} &
 \includegraphics[trim={0 0.5cm 0 0}, clip, width=34mm]{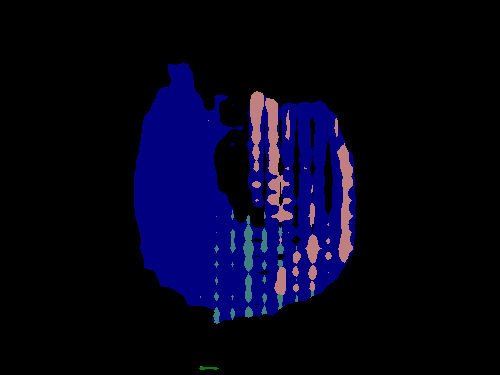} &
 \includegraphics[trim={0 0.5cm 0 0}, clip, width=34mm]{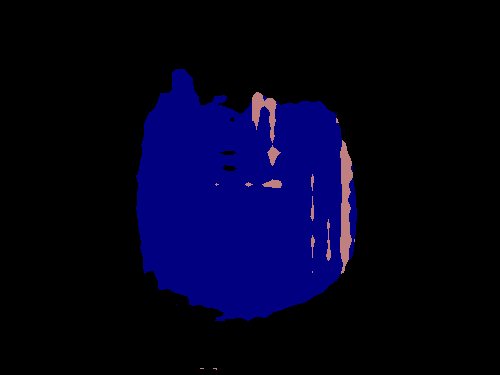}  \\
 
 \includegraphics[trim={0 0 0 0}, clip, width=34mm]{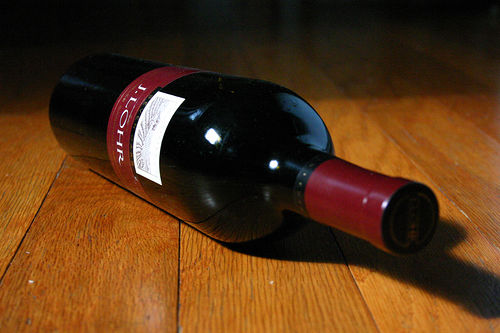} &
 \includegraphics[trim={0 0 0 0}, clip, width=34mm]{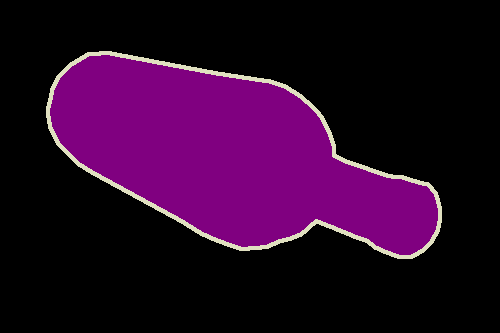} &
 \includegraphics[trim={0 0 0 0}, clip, width=34mm]{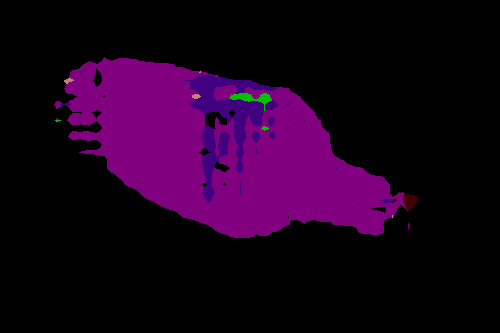} &
 \includegraphics[trim={0 0 0 0}, clip, width=34mm]{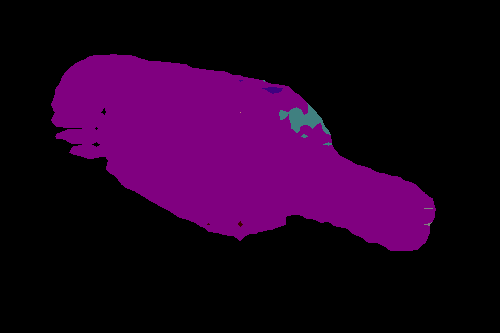} &
 \includegraphics[trim={0 0 0 0}, clip, width=34mm]{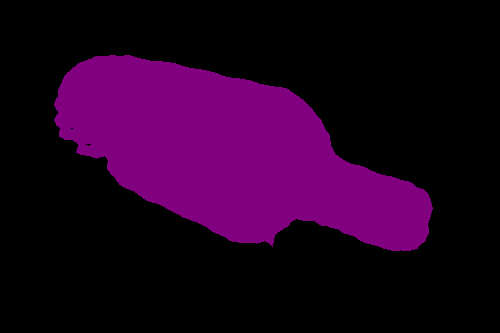}  \\
\end{tabular}
\caption{Qualitative results obtained using our semi-supervised segmentation approach on the PASCAL VOC dataset with 5\% labeled data without COCO pre-training. }
\label{fig:voc_results}
\vspace{-4mm}
\end{figure*}

\subsection{Setup}

\subsubsection{Datasets}

\mypara{PASCAL VOC 2012.} The dataset consists of 20 foreground object classes and one background class. We use the augmented annotation set which consists of 10582 training images and 1449 validation images. The training set contains 1464 images from the original PASCAL data and 9118 extra images from the Segmentation Boundary Dataset (SBD) \cite{BharathICCV2011}.
The training data augmentations include random resizing, cropping to $321 \times 321$, and horizontal flipping. All the results for the PASCAL VOC dataset are shown on the validation set.

\mypara{PASCAL-Context.} This is a whole scene parsing dataset containing 4,998 training and 5,105 testing images with dense semantic labels.
Following the previous work \cite{DBLP:journals/pami/ChenPKMY18, Lin_2017_CVPR, Zhang_2018_CVPR}, we used semantic labels for 60 most frequent classes including the background class. The training data augmentations were the same as for the PASCAL VOC dataset. 

\mypara{Cityscapes.} This is a driving scene dataset with 2975, 500, 1525 densely annotated images for training, validation, and testing, and contains 19 classes.
We downsample the original $1024 \times 2048$ images by a factor 2.
The training data is augmented with random crops of size $256 \times 512$ and horizontal flipping.
All the results on the Cityscapes dataset are shown on the validation set. 

\mypara{Evaluation Metric.} We report mean Intersection-over-Union (mIoU) for all our experiments.

% trim={<left> <lower> <right> <upper>}

\subsubsection{Network Architecture}

% We use 101-layer Deep Residual Network \cite{He_2016_CVPR} (ResNet-101) initialized with pre-trained ImageNet \cite{Deng09imagenet:a} weights, as the backbone network for the generator network of \igan{} branch and student network of the \mt branch. 

\mypara{Semi-supervised Segmentation GAN.} 
We used DeepLabv2 \cite{DBLP:journals/pami/ChenPKMY18} as our main segmentation network.
% We abstain from using multi-scale input fusion as proposed in \cite{DBLP:journals/pami/ChenPKMY18} and use only single-scale input variant of DeepLabv2 due to memory constraints.
Due to memory constraints, we used a single-scale variant of it.
The discriminator network of the GAN model was a standard binary classification network consisting of 4 convolutional layers with $4 \times 4$ kernels with \{64, 128, 256, 512\} channels, each followed by a Leaky-ReLU \cite{Maas13rectifiernonlinearities} activation with negative slope of 0.2 and a dropout \cite{Srivastava:2014:DSW:2627435.2670313} layer with dropout probability of 0.5. We found this high dropout rate to be crucial for stable GAN training. The last convolutional layer is followed by global average pooling and a fully-connected layer. The output vector representation produced after global average pooling is used for evaluating the feature matching loss. 

\mypara{Semi-supervised Multi-label Classification Network.}
We used ResNet-101 \cite{He_2016_CVPR} pre-trained on the ImageNet dataset \cite{Deng09imagenet:a} as the base architecture.

%%---------------------------------------------------------------%%

\subsubsection{Training details} 
 Similar to \cite{DBLP:journals/pami/ChenPKMY18}, we used the poly-learning policy for both the segmentation and the discriminator networks of the GAN model, where the base learning rate was multiplied by a factor of $((1-\frac{\text{iter}}{\text{max\_iter}})^{pow})$ in every iteration. 
 In our setup, $pow=0.9$.
 Following the learning scheme in \cite{Hung_2018_BMVC}, the segmentation network was optimized using the SGD optimizer with a base learning rate of 2.5e-4, momentum 0.9 and a weight decay of 5e-4.
 The discriminator network was optimized using the Adam optimizer \cite{kingma:adam} with a base learning rate of 1e-4 and betas set to $(0.9, 0.99)$. The model was trained for 35K iterations on the PASCAL VOC and Cityscapes dataset, and for 50K iterations on the PASCAL-Context dataset.
 All the learning hyper-parameters remained the same for all datasets except for the Cityscapes dataset, where the base learning rate of the discriminator network was set to 1e-5.
 We used a batch size of 8 for both PASCAL datasets and a batch size of 5 for the Cityscapes dataset. 
 Through cross-validation, we find the optimal loss weights: $\lambda_{fm}= 0.1$, $\lambda_{st}=1.0$, $\lambda_{cons}=1.0$ and $\tau=0.2$. These hyper-parameters remained the same for all datasets, whereas we set
 $\gamma=0.6$ for both PASCAL datasets and $0.7$ for the Cityscapes dataset. 
 %\TODO{Mention exact cross-validation procedure, since you test on the validation set. $\tau$ verified $\gamma, L_{fm}, L_{st} $ in-process}
 Our implementation is based on the open source toolbox Pytorch \cite{paszke2017automatic}.
 All the experiments were run on a Nvidia Tesla P100 GPU.

\subsubsection{Baselines}

We compare to the DeepLabv2 \cite{DBLP:journals/pami/ChenPKMY18} network as the fully-supervised baseline approach, which was trained only on the labeled part of the dataset.
DeepLabv2 makes use of dilated convolutions to enlarge the receptive field size and incorporate larger context, and introduces atrous spatial pyramidal pooling to capture image context at multiple levels.

Our main semi-supervised baseline is the approach proposed by Hung \etal \cite{Hung_2018_BMVC}.

Apart from the differences described in Sec.~\ref{sec:related_work}, they also use a two-stage GAN training. %In the first stage, only $D$ is trained on the labeled data. 
In the first stage, both D and G are trained only using labeled data. In the second stage, $D$'s outputs are used to update $G$ using unlabeled samples, while $D$ itself is further trained only on the labeled images.

%%------------------------------------Results-----------------------------------------------------

\subsection{Results}

\subsubsection{Semi-supervised Semantic Segmentation}

We evaluated our approach with different ratios of labeled and unlabeled samples.
 % In Tables \ref{table:ssl_voc}, \ref{table:ssl_pc}, \ref{table:ssl_city}, column heads (for e.g.\ $1/20, 1/8,...$) refers to the fraction of labeled data used with rest as unlabeled.
$1/50, 1/20, 1/8, 1/4$ are the fractions of the total training images in the dataset that are used as labeled data, the rest of the data was used without labels.
 The labeled samples in the data splits were randomly sampled from the whole dataset, and the same data splits were used for all the baselines.

%%----------------------------PASCAL VOC Dataset----------------------------------------

\begin{threeparttable}[]
\caption{Semi-supervised semantic segmentation results on the PASCAL VOC dataset without and with COCO pre-training.}
    \small
    \setlength\tabcolsep{0pt}
\begin{tabular*}{\linewidth}{@{\extracolsep{\fill}} l cccc @{}}
    \toprule
    \multicolumn{5}{c}{without COCO pre-training} \\
        \cmidrule{1-5}
&   \multicolumn{4}{c}{Labeled Data} \\
        \cmidrule{2-5}
Method                          & 1/50  
                                                & 1/20
                                                                & 1/8
                                                                                & Full\\
        \midrule
DeepLabv2                       & 48.3          & 56.8          & 62.0          & 70.7\\
Hung \etal \cite{Hung_2018_BMVC}& 49.2          & 59.1          & 64.3          & 71.4\\
Ours (\igan{} only)             & 58.1          & 60.9          & 65.4          & 71.2\\
Ours (\igan{} + MLMT)           & \textbf{60.4}  & \textbf{62.9} & \textbf{67.3} & \textbf{73.2}\\

        \cmidrule{1-5}
    \multicolumn{5}{c}{with COCO pre-training} \\
        \cmidrule{1-5}

DeepLabv2                       & 53.2          & 58.7          & 65.2          & 73.6\\
Hung \etal \cite{Hung_2018_BMVC}& 57.2          & 64.7          & 69.5          & 74.9\\
Ours (\igan{} only)             & 60.9          & 66.4          & 69.8          & 73.9\\
Ours (\igan{} + MLMT)           & \textbf{63.3}  & \textbf{67.2} & \textbf{71.4} & \textbf{75.6}\\
        \bottomrule
\end{tabular*}

\label{table:ssl_voc}
\end{threeparttable} \\

\begin{figure}
\begin{tabular}{c@{\hspace{1mm}}c@{\hspace{1mm}}c@{\hspace{1mm}}c}
Original & Ground truth & Baseline  & Ours  \\[6pt]

\includegraphics[trim={0 0cm 0 1cm}, clip,width=21mm]{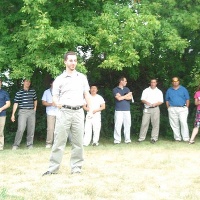} &
\includegraphics[trim={0 0cm 0 1cm}, clip,width=21mm]{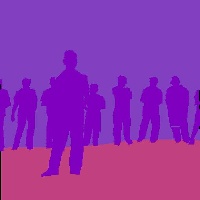} &
\includegraphics[trim={0 0cm 0 1cm}, clip,width=21mm]{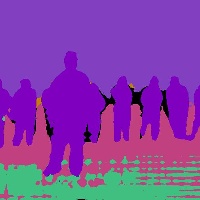} &
\includegraphics[trim={0 0cm 0 1cm}, clip,width=21mm]{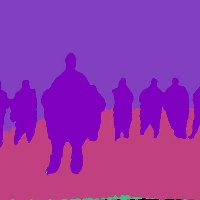} \\
\includegraphics[width=21mm]{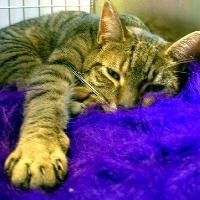} &
\includegraphics[width=21mm]{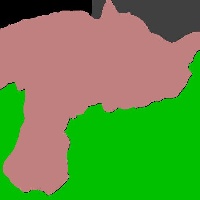} &
\includegraphics[width=21mm]{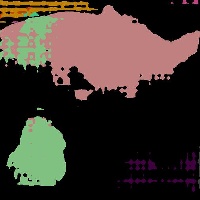} &
\includegraphics[width=21mm]{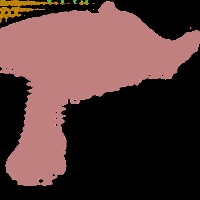} \\
\includegraphics[width=21mm]{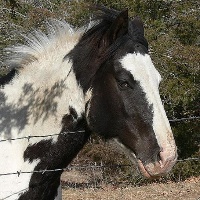} &
\includegraphics[width=21mm]{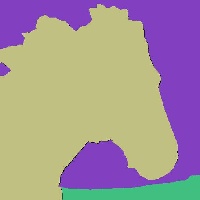} &
\includegraphics[width=21mm]{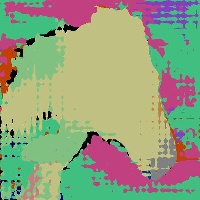} &
\includegraphics[width=21mm]{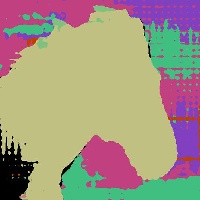} \\
\includegraphics[trim={0 0cm 0 1cm}, clip, width=21mm]{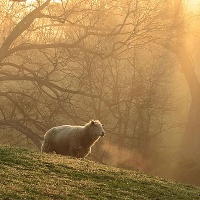} &
\includegraphics[trim={0 0cm 0 1cm}, clip, width=21mm]{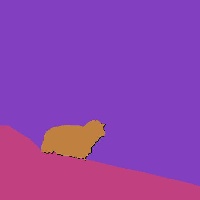} &
\includegraphics[trim={0 0cm 0 1cm}, clip, width=21mm]{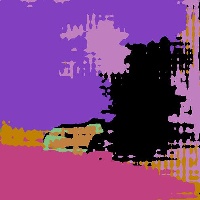} &
\includegraphics[trim={0 0cm 0 1cm}, clip, width=21mm]{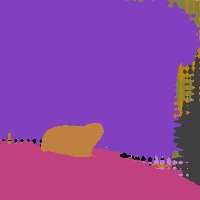} \\
\includegraphics[width=21mm]{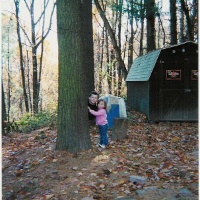} &
\includegraphics[width=21mm]{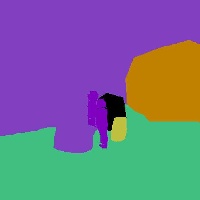} &
\includegraphics[width=21mm]{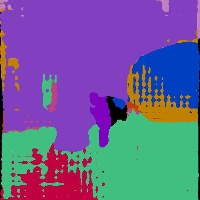} &
\includegraphics[width=21mm]{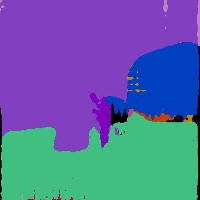} \\
\includegraphics[trim={0 0cm 0 1cm}, clip,width=21mm]{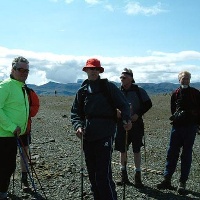} &
\includegraphics[trim={0 0cm 0 1cm}, clip,width=21mm]{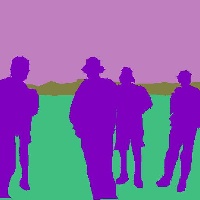} &
\includegraphics[trim={0 0cm 0 1cm}, clip,width=21mm]{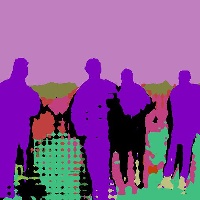} &
\includegraphics[trim={0 0cm 0 1cm}, clip,width=21mm]{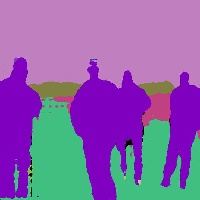} \\
\includegraphics[width=21mm]{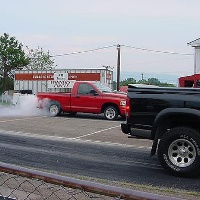} &
\includegraphics[width=21mm]{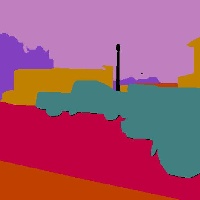} &
\includegraphics[width=21mm]{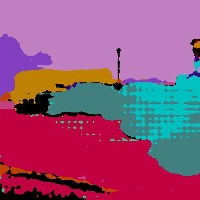} &
\includegraphics[width=21mm]{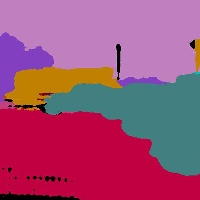} \\

\includegraphics[trim={0 0cm 0 1cm}, clip,width=21mm]{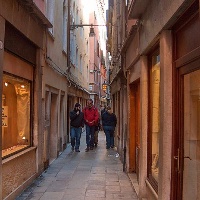} &
\includegraphics[trim={0 0cm 0 1cm}, clip,width=21mm]{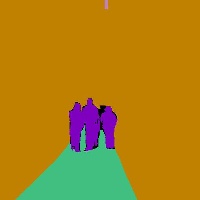} &
\includegraphics[trim={0 0cm 0 1cm}, clip,width=21mm]{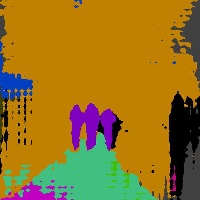} &
\includegraphics[trim={0 0cm 0 1cm}, clip,width=21mm]{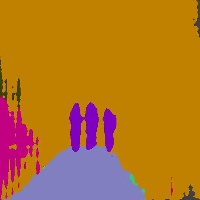} \\
\includegraphics[width=21mm]{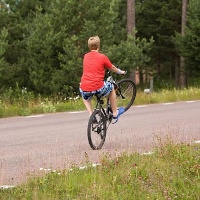} &
\includegraphics[width=21mm]{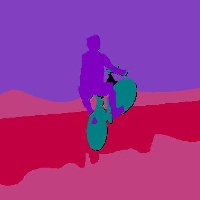} &
\includegraphics[width=21mm]{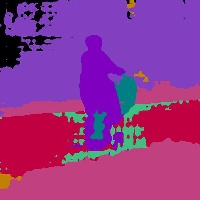} &
\includegraphics[width=21mm]{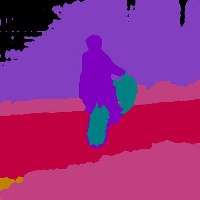} \\
\includegraphics[width=21mm]{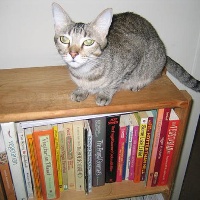} &
\includegraphics[width=21mm]{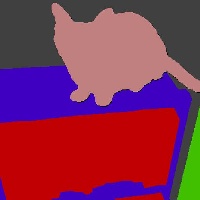} &
\includegraphics[width=21mm]{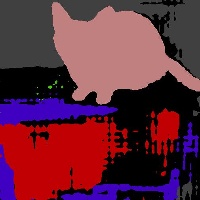} &
\includegraphics[width=21mm]{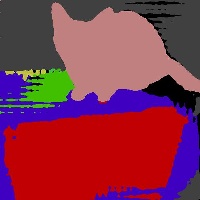} \\

\end{tabular}
\caption{Qualitative results on the PASCAL-Context dataset using 1/8 labeled samples. Our approach produces improved results compared to the baseline. We compare our ('Ours') results with the fully-supervised baseline which is trained only on the labeled subset of data.}
\label{fig:sup_pc1}
\end{figure}

\mypara{PASCAL VOC Dataset.}
Table \ref{table:ssl_voc} shows the segmentation results on the PASCAL VOC dataset with and without pre-training on the Microsoft COCO \cite{mscoco} dataset.
We achieve improved results compared to the previous method for all data splits. Our method achieves a performance increase of 5\% to 12\% over the baseline for different data splits by utilizing unlabeled samples without pre-training the network on any segmentation dataset.
Notably, the approach works well even with only 2\% (1/50) of labeled data.
Figure \ref{fig:voc_results} shows qualitatively how our method helps remove artifacts produced by other methods.
We also validated our approach with COCO pre-training to directly compare with Hung \etal \cite{Hung_2018_BMVC}, and achieved an improvement of 6.1 mIoU points over them for the 1/50 split.
We speculate that \cite{Hung_2018_BMVC} is inferior in the low-data regime due to the two-stage GAN training, where the discriminator is only updated based on the labeled samples. This effectively reduces the amount of data it sees during training, which can easily lead to overfitting.

We conducted our initial experiments using Deeplabv3+ as the backbone architecture.
Deeplabv3+ is unstable in the low-data `supervised only' setting. It is only superior, if there is much labeled data. Thus, for a more informative experiment, we rather used Deeplabv2. However, our semi-supervised model achieves even better performance with Deeplabv3+ than with the DeepLabv2-based model, see Table \ref{table:ssl_v2_v3p}.

\begin{threeparttable}[htb]
\caption{Results on PASCAL VOC without COCO pre-training using different backbone architectures. }
    \small
    \setlength\tabcolsep{0pt}
\begin{tabular*}{\linewidth}{@{\extracolsep{\fill}} l cccc @{}}
    \toprule
%&   \multicolumn{2}{c}{Labeled Data} \\
%        \cmidrule{2-3}
Method                          & 1/50  
                                            & 1/20 
                                                        & 1/8
                                                                    & Full\\
        \midrule
Deeplabv2 (v2)                  & 48.3       & 56.8     & 62.0      & 70.7\\
%igan                           & 53.9       & 61.2 \\
Ours v2 (\igan + \mt)           & 60.4       & 62.9     & 67.3      & 73.2\\
Deeplabv3+ (v3+)                & unstable   & unstable  & 63.5      & 74.6 \\
%v3+ + \igan                    & 62.4       & 66.5 \\
Ours v3+ (\igan + \mt)          & 62.6       & 66.6     & 70.4      & 74.7\\

        \bottomrule
\end{tabular*}

\label{table:ssl_v2_v3p}
\end{threeparttable} \\

The results were obtained with cross-validation to avoid hyper-parameter search on the evaluation set. We also submitted our results to the PASCAL test server. Due to the benchmark restrictions we could only submit one random split (5\% labeled samples). The results are consistent with our previous conclusions: 50.1 mIoU for baseline DeepLabv2 vs 60.5 for our semi-supervised method. 

\begin{figure}
\centering
\begin{tabular}{c@{\hspace{1mm}}c@{\hspace{1mm}}c@{\hspace{1mm}}c}
%(a) Original & (b) Ground Truth & (c) Baseline  & (d) Our Results  \\[6pt]

\begin{turn}{90}\hspace{7mm}Orig\end{turn} &
\includegraphics[width=35mm]{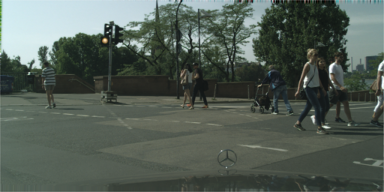} &
\begin{turn}{90}\hspace{7mm}GT\end{turn} &
\includegraphics[width=35mm]{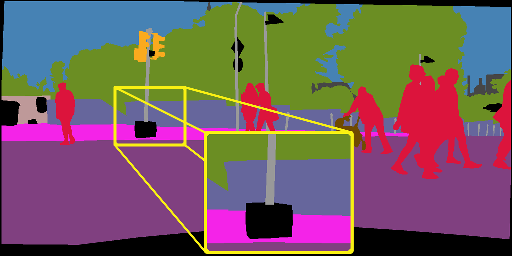} \\
% \vspace{2mm}
\begin{turn}{90}\hspace{7mm}Base\end{turn} &
\includegraphics[width=35mm]{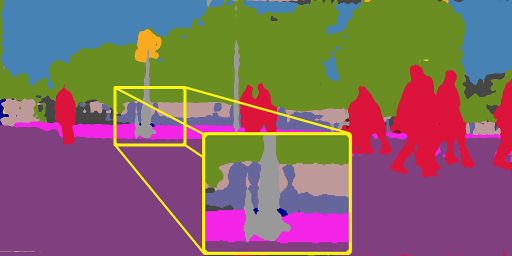} &
\begin{turn}{90}\hspace{7mm}Ours\end{turn} &
\includegraphics[width=35mm]{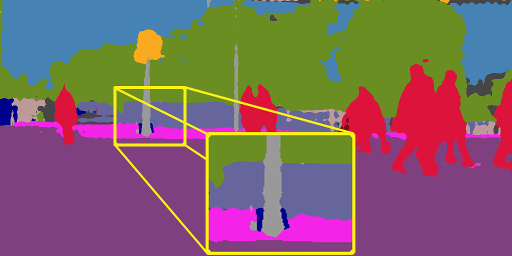} \\

\begin{turn}{90}\hspace{7mm}Orig\end{turn} &
\includegraphics[width=35mm]{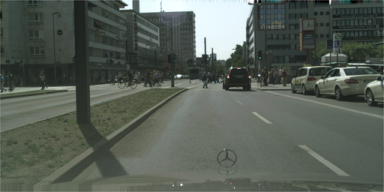} &
\begin{turn}{90}\hspace{7mm}GT\end{turn} &
\includegraphics[width=35mm]{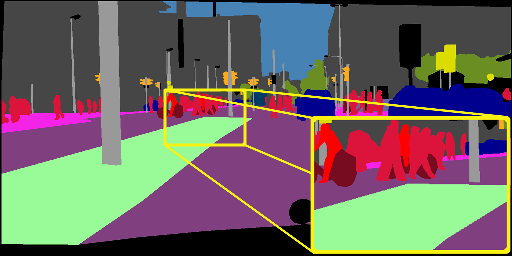} \\
% \vspace{2mm}
\begin{turn}{90}\hspace{7mm}Base\end{turn} &
\includegraphics[width=35mm]{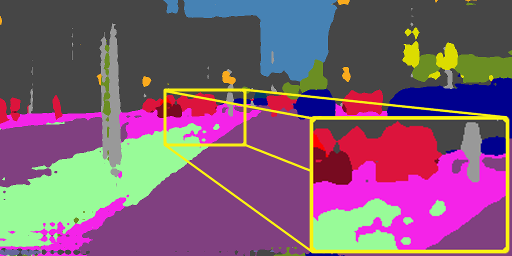} &
\begin{turn}{90}\hspace{7mm}Ours\end{turn} &
\includegraphics[width=35mm]{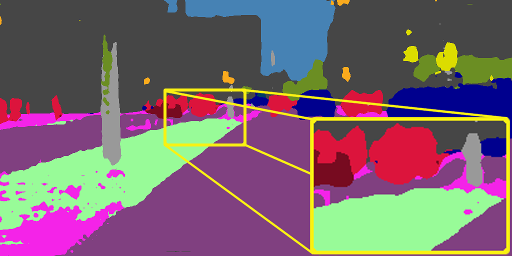} \\

\begin{turn}{90}\hspace{7mm}Orig\end{turn} &
\includegraphics[width=35mm]{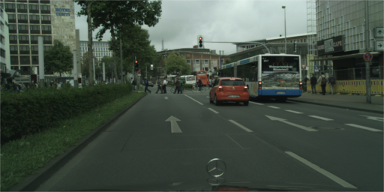} &
\begin{turn}{90}\hspace{7mm}GT\end{turn} &
\includegraphics[width=35mm]{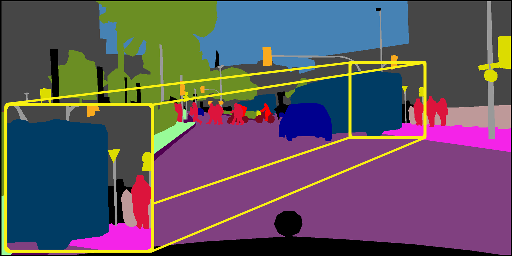} \\
% \vspace{2mm}
\begin{turn}{90}\hspace{7mm}Base\end{turn} &
\includegraphics[width=35mm]{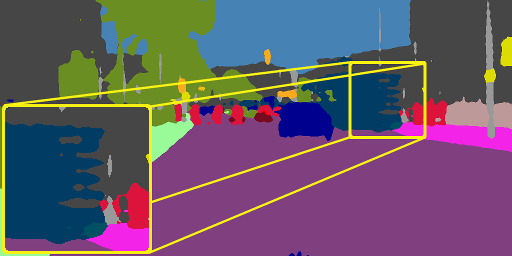} &
\begin{turn}{90}\hspace{7mm}Ours\end{turn} &
\includegraphics[width=35mm]{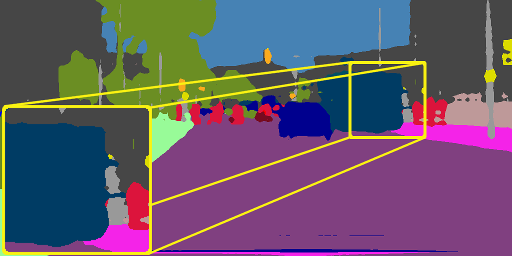} \\

 \begin{turn}{90}\hspace{7mm}Orig\end{turn} &
\includegraphics[width=35mm]{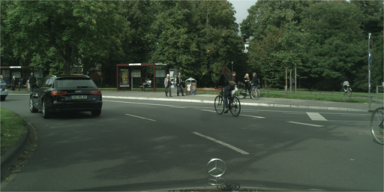} &
\begin{turn}{90}\hspace{7mm}GT\end{turn} &
\includegraphics[width=35mm]{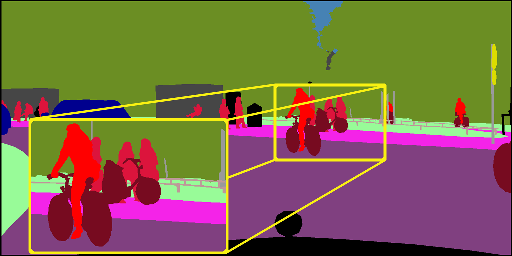} \\
% \vspace{2mm}
\begin{turn}{90}\hspace{7mm}Base\end{turn} &
\includegraphics[width=35mm]{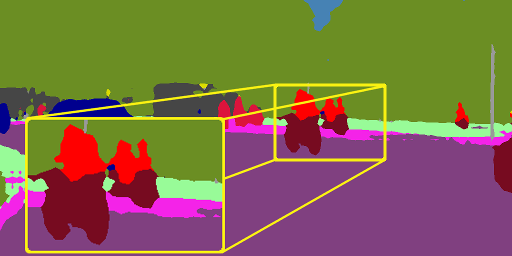} &
\begin{turn}{90}\hspace{7mm}Ours\end{turn} &
\includegraphics[width=35mm]{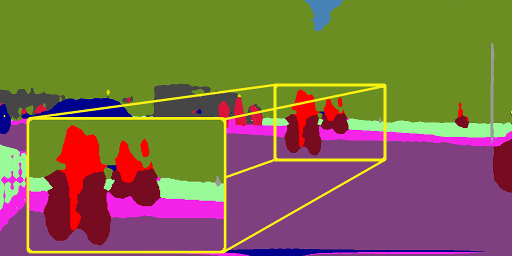} \\

\begin{turn}{90}\hspace{7mm}Orig\end{turn} &
\includegraphics[width=35mm]{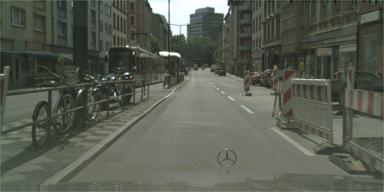} &
\begin{turn}{90}\hspace{7mm}GT\end{turn} &
\includegraphics[width=35mm]{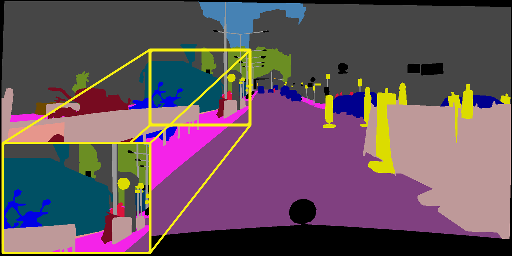} \\
% \vspace{2mm}
\begin{turn}{90}\hspace{7mm}Base\end{turn} &
\includegraphics[width=35mm]{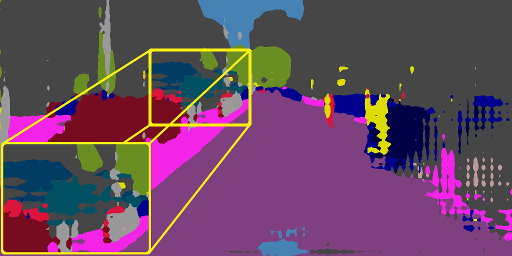} &
\begin{turn}{90}\hspace{7mm}Ours\end{turn} &
\includegraphics[width=35mm]{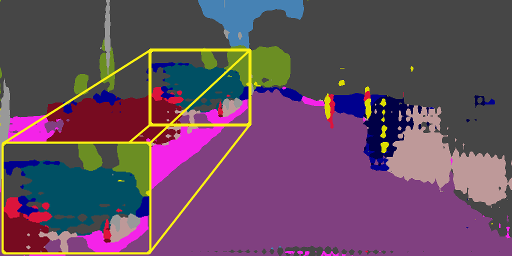} \\

\begin{turn}{90}\hspace{7mm}Orig\end{turn} &
\includegraphics[width=35mm]{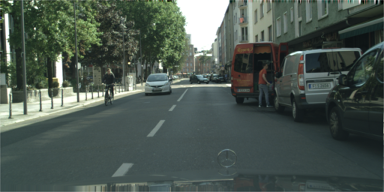} &
\begin{turn}{90}\hspace{7mm}GT\end{turn} &
\includegraphics[width=35mm]{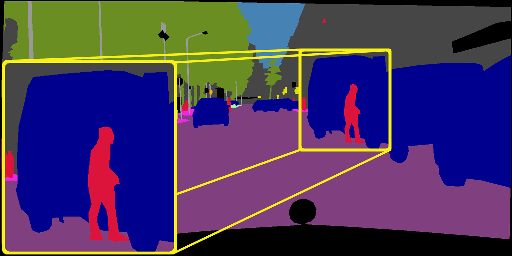} \\
% \vspace{2mm}
\begin{turn}{90}\hspace{7mm}Base\end{turn} &
\includegraphics[width=35mm]{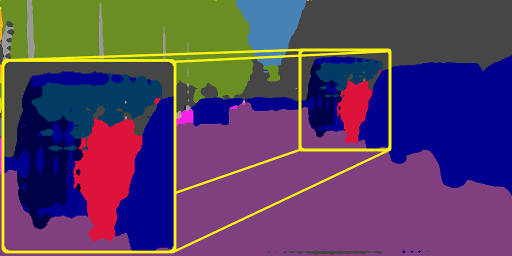} &
\begin{turn}{90}\hspace{7mm}Ours\end{turn} &
\includegraphics[width=35mm]{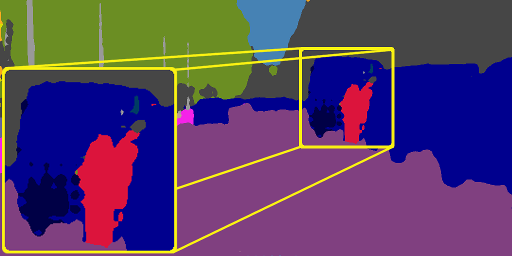} \\

\end{tabular}
\caption{Qualitative results on the Cityscapes dataset using 1/8 labeled samples without COCO pre-training. The proposed semi-supervised approach produces improved results compared to the baseline. We compare our ('Ours') results with the fully-supervised baseline (`Base') which is trained only on the labeled subset of data.}
\label{fig:sup_city1}
\end{figure}

%%----------------------------PASCAL-Context Dataset------------------------------------------

\mypara{PASCAL-Context Dataset.}
Our approach successfully generalizes to the whole scene parsing PASCAL-Context dataset. Table \ref{table:ssl_pc} shows the performance on two splits (1/8 and 1/4 labeled data) of PASCAL-Context.
Although this dataset is smaller and more difficult than PASCAL VOC, there is still an improvement over the baseline of 3.2\% and 2.4\% for the 1/8 and 1/4 splits, respectively. 

Fig.~\ref{fig:sup_pc1} show qualitative results on the PASCAL-Context test set using 1/8 labeled samples and the remaining unlabeled samples. PASCAL-Context is a smaller and harder dataset as compared to PASCAL VOC, therefore the results are not as visually appealing. Still, there is a clear improvement over the baseline.

\begin{threeparttable}[]
\caption{Semi-supervised semantic segmentation results on the PASCAL-Context dataset without COCO pre-training.}
    \small
    \setlength\tabcolsep{0pt}
\begin{tabular*}{\linewidth}{@{\extracolsep{\fill}} l ccc @{}}
    \toprule
&   \multicolumn{3}{c}{Labeled Data} \\
        \cmidrule{2-4}
Method                          & 1/8
                                                & 1/4
                                                                & Full\\ 
        \midrule
DeepLabv2                   & 32.1          & 35.4          & 41.0\\
Hung \etal \cite{Hung_2018_BMVC} & 32.8     & 34.8          & 39.1\\
Ours (\igan{} only)         & 34.4          & 37.1          & 40.8\\
Ours (\igan{} + MLMT)       & \textbf{35.3} & \textbf{37.8} & \textbf{41.1}\\
        \bottomrule
\end{tabular*}

\label{table:ssl_pc}
\begin{tablenotes}[para,flushleft,small]
\end{tablenotes}
\end{threeparttable} \\

\begin{figure*}[h!]
\centering
\begin{tabular}{c@{\hspace{1mm}}c@{\hspace{1mm}}c@{\hspace{1mm}}c@{\hspace{1mm}}c@{\hspace{1mm}}c@{\hspace{1mm}}c}
(a) Original & (b) GT & (c) Baseline & (d) \mt{}  & (e) \igan{}  & (f)\igan{}+CNN & (g)\igan{}+\mt{} \\[6pt]
 \includegraphics[trim={1cm 0 0 0}, clip, width=24mm]{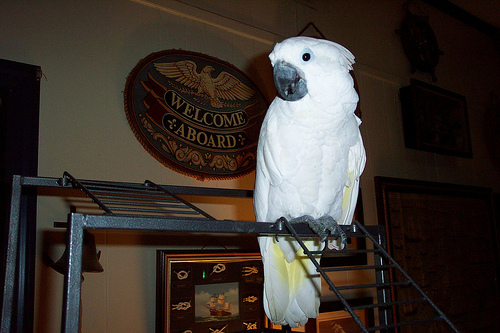} &
 \includegraphics[trim={1cm 0 0 0}, clip, width=24mm]{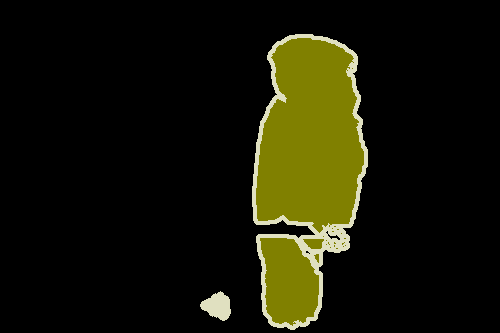} &
 \includegraphics[trim={1cm 0 0 0}, clip, width=24mm]{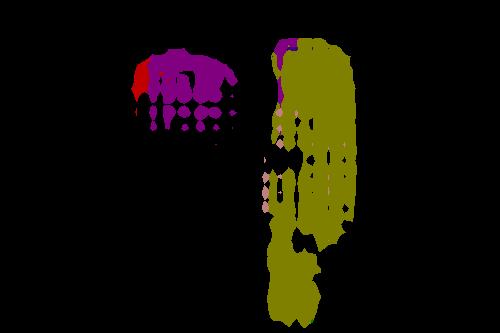} &
 \includegraphics[trim={1cm 0 0 0}, clip, width=24mm]{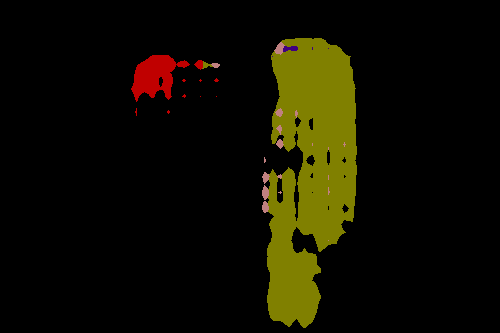} &
 \includegraphics[trim={1cm 0 0 0}, clip, width=24mm]{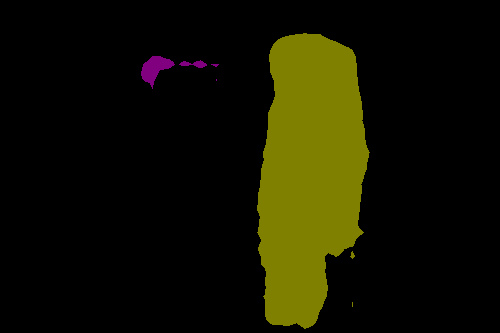} &
 \includegraphics[trim={1cm 0 0 0}, clip, width=24mm]{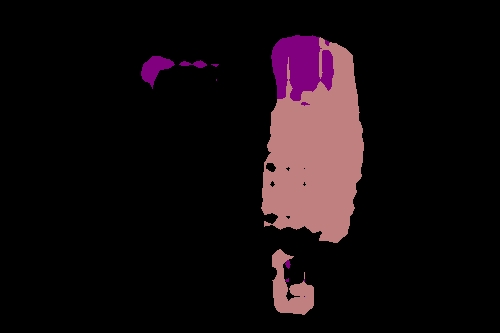} &
 \includegraphics[trim={1cm 0 0 0}, clip, width=24mm]{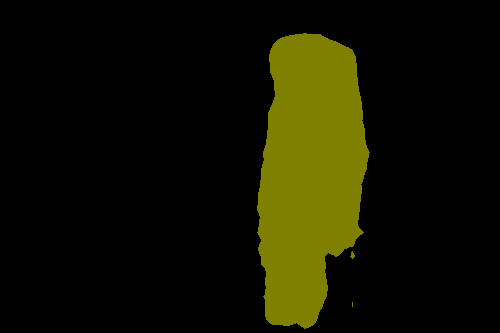} \\
 
 \includegraphics[trim={1cm 0 0 0}, clip, width=24mm]{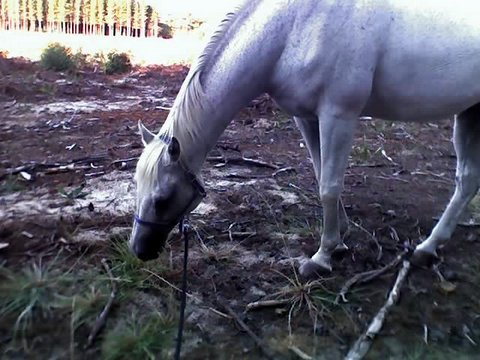} &
 \includegraphics[trim={1cm 0 0 0}, clip, width=24mm]{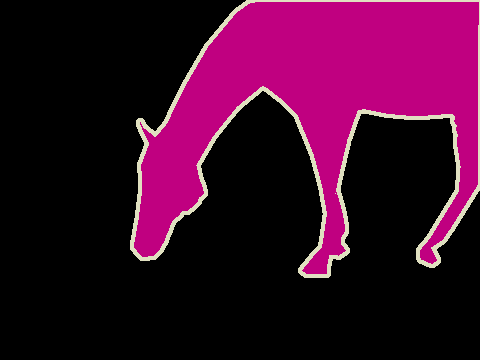} &
 \includegraphics[trim={1cm 0 0 0}, clip, width=24mm]{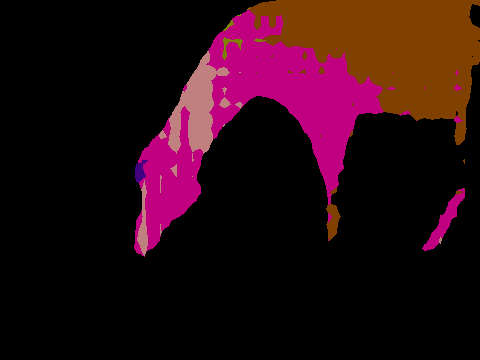} &
 \includegraphics[trim={1cm 0 0 0}, clip, width=24mm]{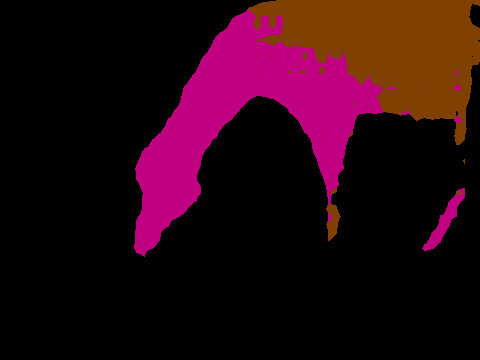} &
 \includegraphics[trim={1cm 0 0 0}, clip, width=24mm]{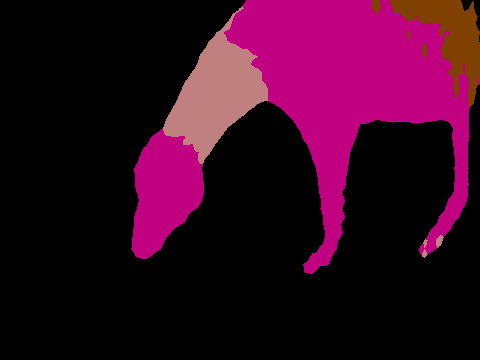} &
 \includegraphics[trim={1cm 0 0 0}, clip, width=24mm]{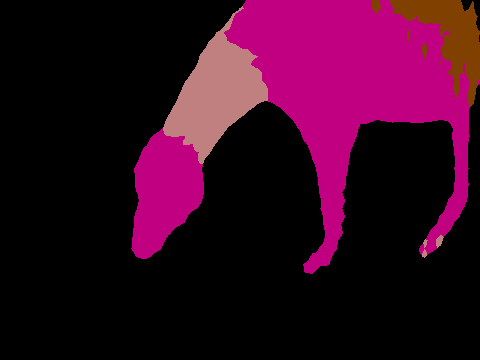} &
 \includegraphics[trim={1cm 0 0 0}, clip, width=24mm]{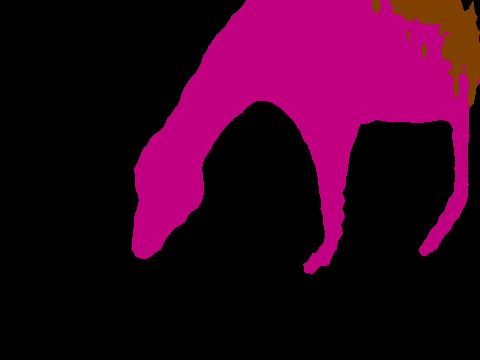} \\
 
 \includegraphics[trim={0 1cm 0 1cm}, clip, width=24mm]{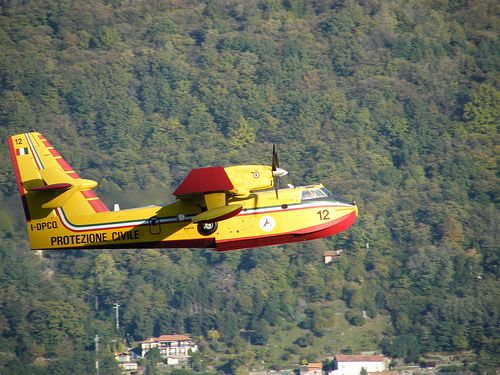} &
 \includegraphics[trim={0 1cm 0 1cm}, clip, width=24mm]{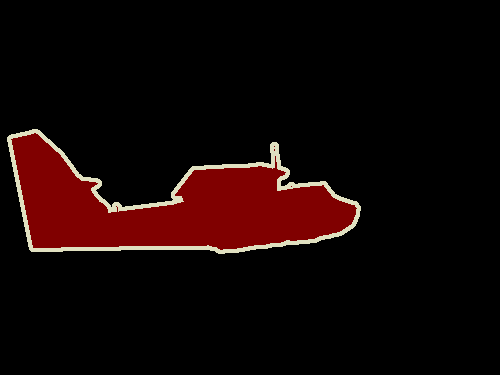} &
 \includegraphics[trim={0 1cm 0 1cm}, clip, width=24mm]{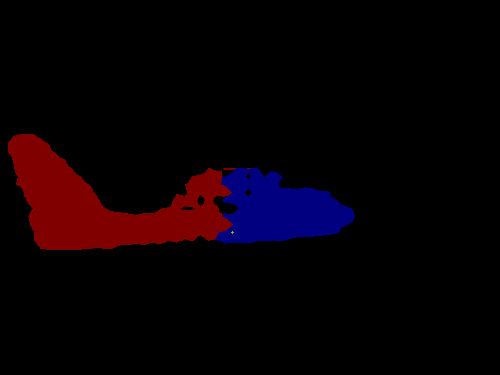} &
 \includegraphics[trim={0 1cm 0 1cm}, clip, width=24mm]{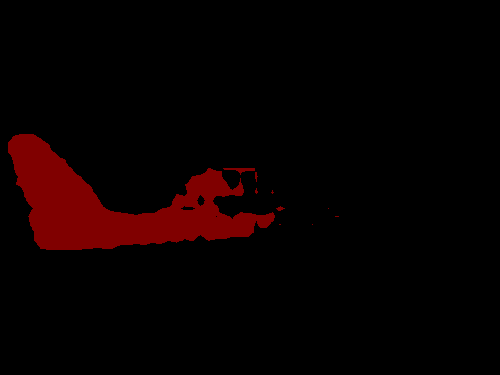} &
 \includegraphics[trim={0 1cm 0 1cm}, clip, width=24mm]{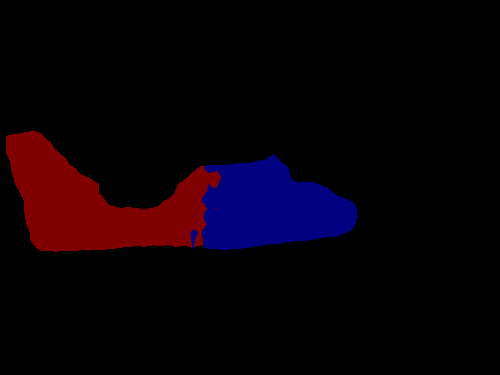} &
 \includegraphics[trim={0 1cm 0 1cm}, clip, width=24mm]{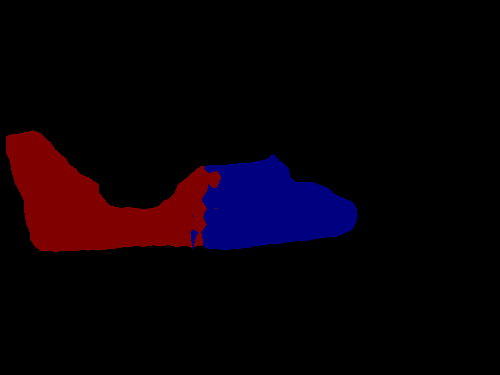} &
 \includegraphics[trim={0 1cm 0 1cm}, clip, width=24mm]{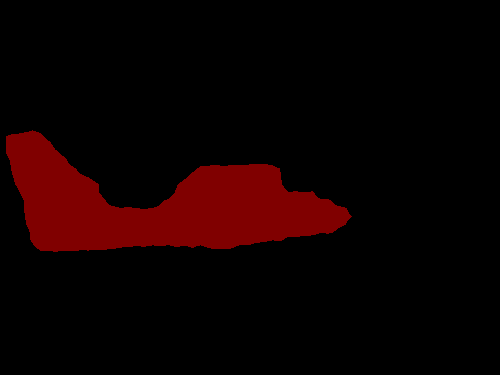} \\
 
 \includegraphics[trim={0 0 0 0}, clip, width=24mm]{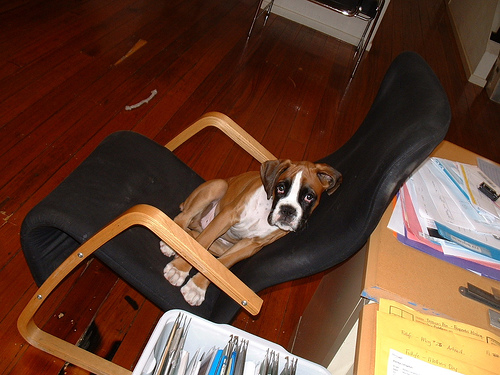} &
 \includegraphics[trim={0 0 0 0}, clip, width=24mm]{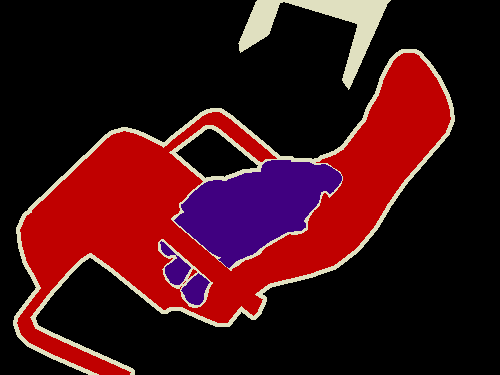} &
 \includegraphics[trim={0 0 0 0}, clip, width=24mm]{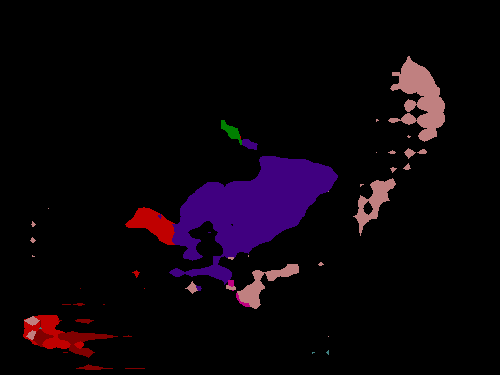} &
 \includegraphics[trim={0 0 0 0}, clip, width=24mm]{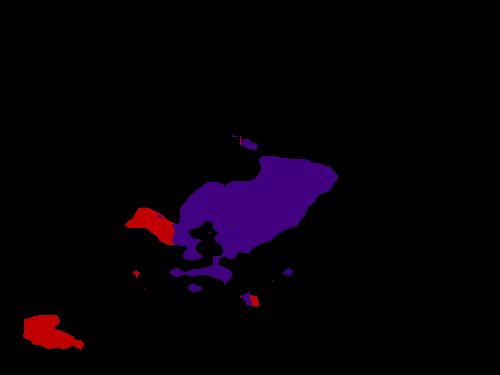} &
 \includegraphics[trim={0 0 0 0}, clip, width=24mm]{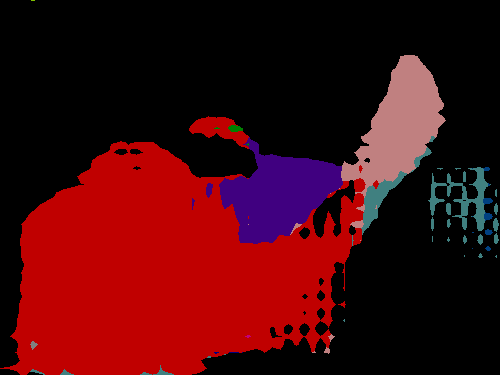} &
 \includegraphics[trim={0 0 0 0}, clip, width=24mm]{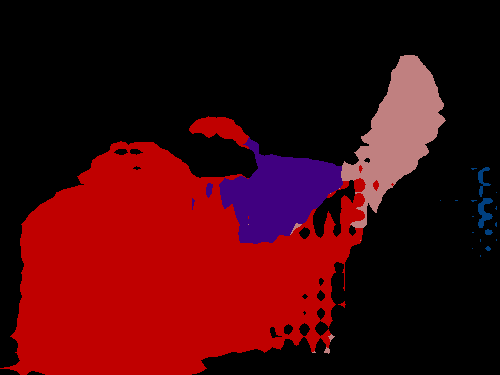} &
 \includegraphics[trim={0 0 0 0}, clip, width=24mm]{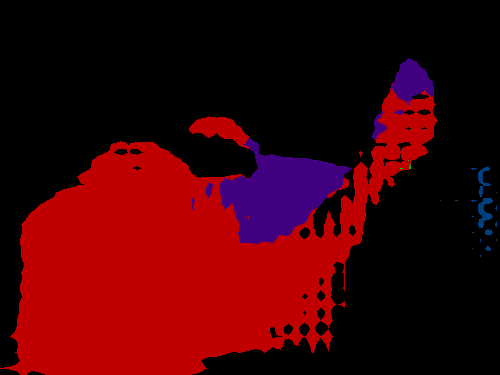} \\

 \includegraphics[trim={0 0 0 0}, clip, width=24mm]{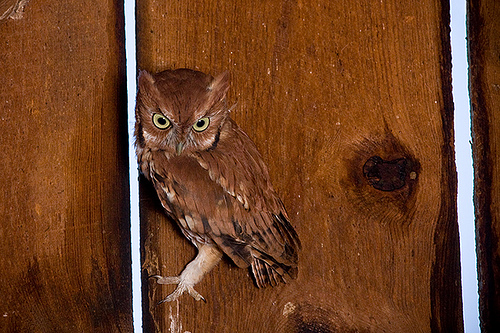} &
 \includegraphics[trim={0 0 0 0}, clip, width=24mm]{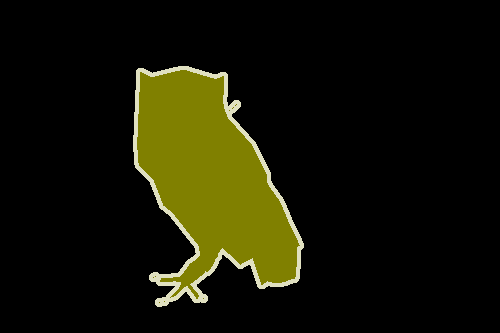} &
 \includegraphics[trim={0 0 0 0}, clip, width=24mm]{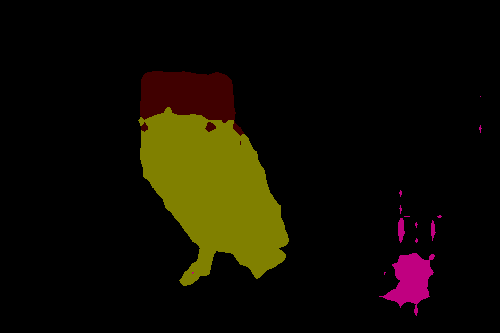} &
 \includegraphics[trim={0 0 0 0}, clip, width=24mm]{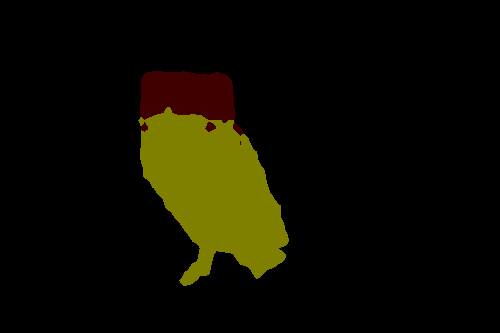} &
 \includegraphics[trim={0 0 0 0}, clip, width=24mm]{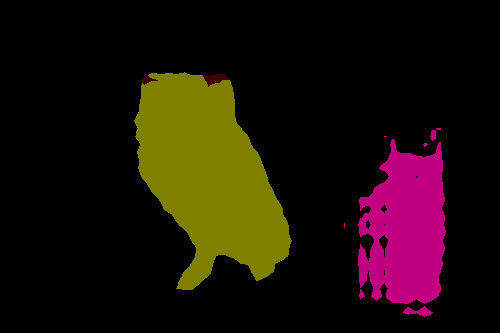} &
 \includegraphics[trim={0 0 0 0}, clip, width=24mm]{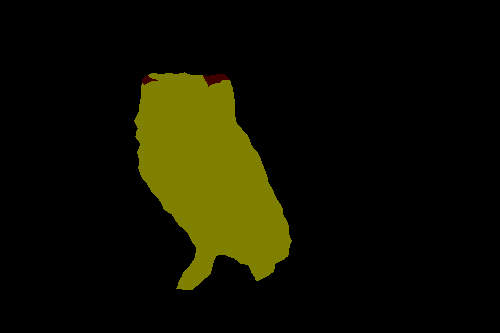} &
 \includegraphics[trim={0 0 0 0}, clip, width=24mm]{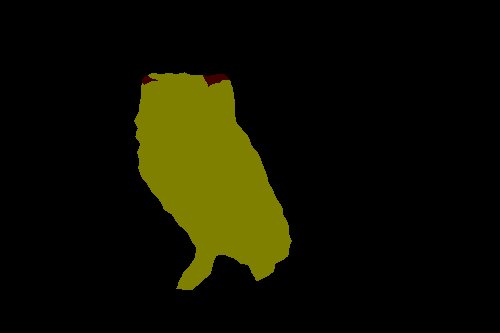} 
 
\end{tabular}
\caption{Ablation study on the PASCAL VOC dataset showing the contribution of the \mt{} (d) and the \igan{} (e) branches individually. The \igan{} and the \mt{} branches together show a complementary behaviour fixing both low and high-level artifacts (g). These results are obtained using 5\% labeled data.}
\label{fig:ab_voc}

\end{figure*}

%%------------------Cityscapes Dataset------------------------

\mypara{Cityscapes Dataset.}
%For the Cityscapes dataset, given only a few labeled images, it is quite challenging to learn from unlabeled samples.
On the Cityscapes dataset, the \igan{} branch yields an improvement over the baseline of 3.1\% and 1.7\% for the 1/8 and 1/4 data splits respectively; see Table \ref{table:ssl_city}.
The distribution of different classes in this dataset is highly imbalanced. 
The vast majority of the classes are present in almost every image, and the few remaining classes occur only scarcely.
In this situation, a classifier that eliminates labels of non-existing classes does not help, thus, our MLMT branch was ineffective for the Cityscapes dataset.

\begin{threeparttable}[]
\caption{Semi-supervised semantic segmentation results on the Cityscapes dataset without COCO pre-training.}

    \small
    \setlength\tabcolsep{0pt}
    
\begin{tabular*}{\linewidth}{@{\extracolsep{\fill}} l ccc @{}}
    \toprule
&   \multicolumn{3}{c}{Labeled Data} \\
        \cmidrule{2-4}
Method                                      & 1/8
                                                                & 1/4
                                                                                & Full\\ 
        \midrule
DeepLabv2                                   & 56.2              & 60.2          & 66.0\\
%Hung \etal \cite{Hung_2018_BMVC} \tnote{a}  & 56.1              & 59.4          & ---\\
Hung \etal \cite{Hung_2018_BMVC}            & 57.1              & 60.5          & \textbf{66.2}\\
Ours (\igan{} only)                         & \textbf{59.3}     & \textbf{61.9} & 65.8\\
%\textbf{\color{red}Ours \igan{} + \mt}      & {\color{red}59.3} & \textbf{61.9} & ToDo\\ with 0.03 threshold
        \bottomrule
\end{tabular*}

\label{table:ssl_city}
\end{threeparttable}\\

Fig.~\ref{fig:sup_city1} show qualitative results obtained using our approach with 1/8 labeled samples and the remaining unlabeled samples. The differences on the Cityscapes dataset are subtle, therefore we include the zoomed-in views of informative areas. On images from Fig.~\ref{fig:sup_city1} show our approach yields improvement over the baseline.

\begin{figure}[h!]
\centering
\begin{tabular}{@{}c@{\hspace{1mm}}c@{\hspace{1mm}}c@{\hspace{1mm}}c@{}}
(a) Original & (b) GT & (c) Baseline & (d) Ours    \\[6pt]
\includegraphics[width=0.24\linewidth]{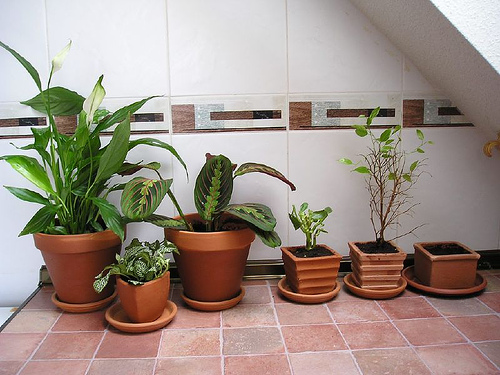} &
\includegraphics[width=0.24\linewidth]{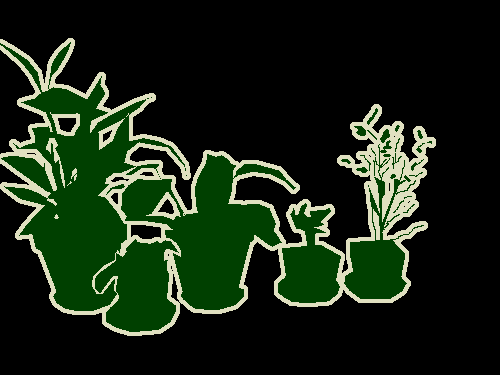} &
\includegraphics[width=0.24\linewidth]{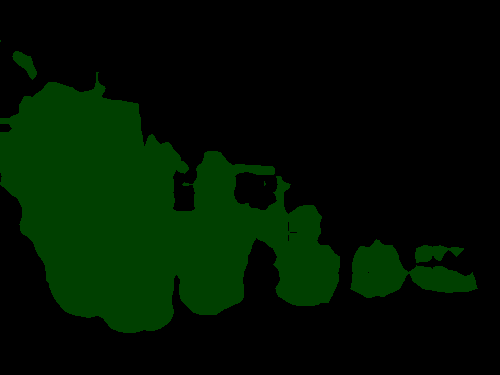} &
\includegraphics[width=0.24\linewidth]{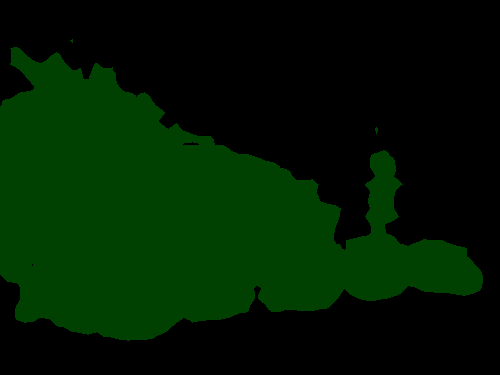}  \\
\includegraphics[trim={0 0 0 1.5cm}, clip, width=0.24\linewidth]{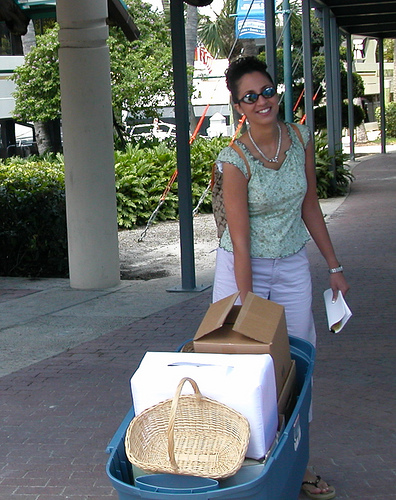} &
\includegraphics[trim={0 0 0 1.5cm}, clip, width=0.24\linewidth]{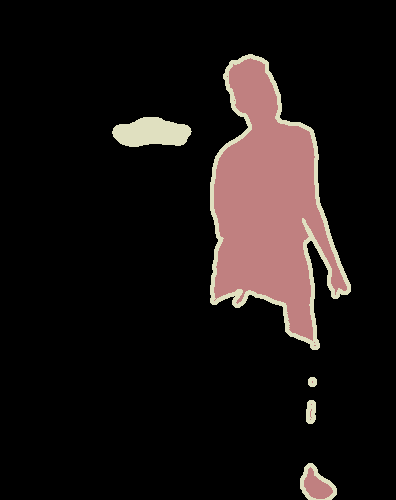} &
\includegraphics[trim={0 0 0 1.5cm}, clip, width=0.24\linewidth]{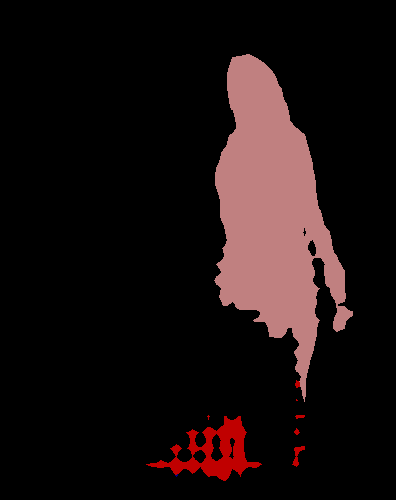} &
\includegraphics[trim={0 0 0 1.5cm}, clip, width=0.24\linewidth]{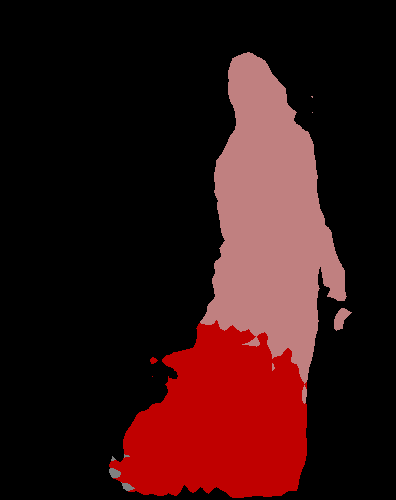}  
\end{tabular}
\caption{Failure cases. Sometimes, our approach can lead to under-segmentation of objects with multiple protrusions, shown in the top row. Bottom row shows a case where an ambiguous foreground object is falsely marked as one of the classes.}% (`chair' in this case).}
\label{fig:fail}
\end{figure}

%%--------------------------------Ablation Studies----------------------------------
\subsubsection{Ablation Studies} \label{sec:ab}
All the experiments for the ablation studies are shown on the PASCAL VOC dataset without COCO pre-training. 

%%-------------------------------------Study-I-------------------------------------
\textbf{Contribution of the two branches.}

Table \ref{table:ab_branch} shows the contribution of the \igan{} branch and the MLMT branch.
The \igan{} branch is able to extract extra dense information using unlabeled images. It improves the shape of the segmented objects, makes the segmentation prediction more coherent by filling small holes, and improves the fine boundaries between the foreground and background.
We showcase these improvements in Figure \ref{fig:ab_voc}(e).

The MLMT branch plays a complementary role and removes the false positives from the predictions. Figure \ref{fig:ab_voc}(d) shows the improvement using the `\mt{} branch only' with the segmentation baseline method and Figure \ref{fig:ab_voc}(g) shows the improvement using the \mt{} branch together with the \igan{} branch. The \mt{} branch makes use of unlabeled images to extract image-level information about the presence of the certain classes in the image. For some cases, the \igan{} branch introduces new artifacts which are also filtered out by the \mt{} branch. This effect is shown in the bottom-row example of Figure \ref{fig:ab_voc}. 

\begin{figure}
\begin{tabular}{c@{\hspace{1mm}}c@{\hspace{1mm}}c@{\hspace{1mm}}c}
Original & Ground truth & Baseline  & Ours  \\[6pt]

\includegraphics[width=21mm]{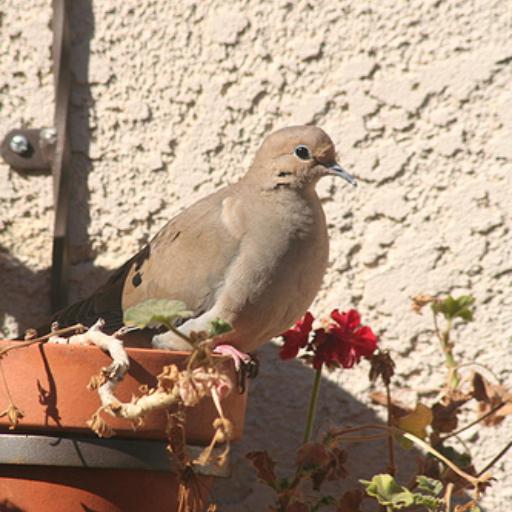} &
\includegraphics[width=21mm]{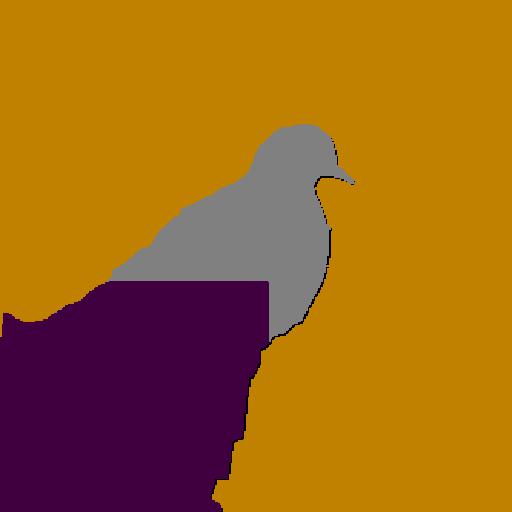} &
\includegraphics[width=21mm]{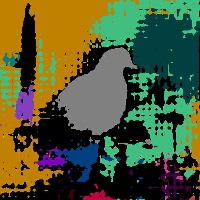} &
\includegraphics[width=21mm]{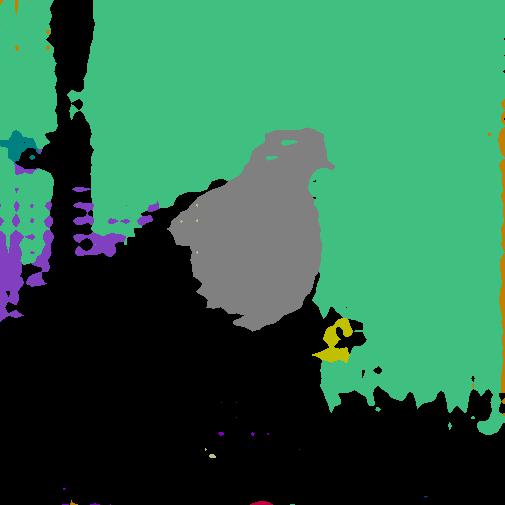} \\
\includegraphics[width=21mm]{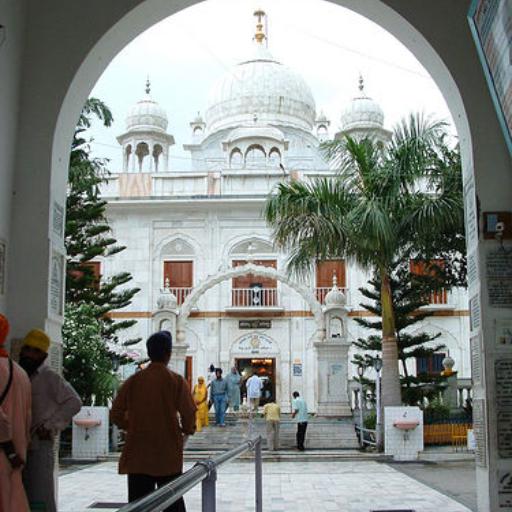} &
\includegraphics[width=21mm]{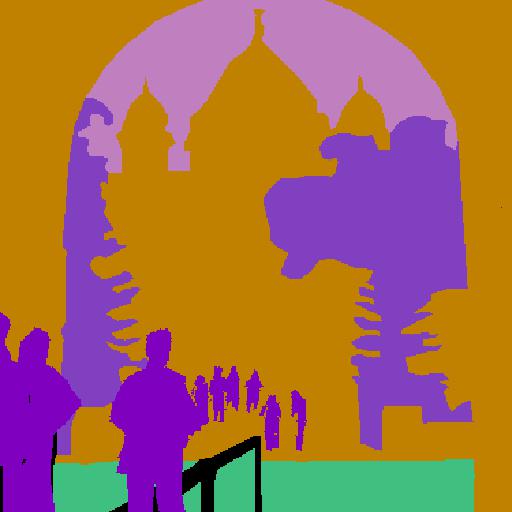} &
\includegraphics[width=21mm]{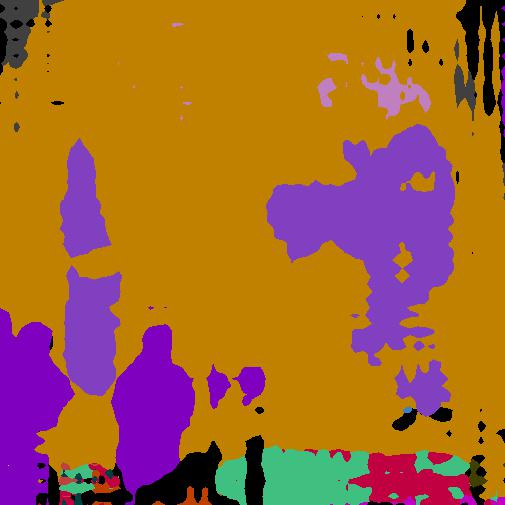} &
\includegraphics[width=21mm]{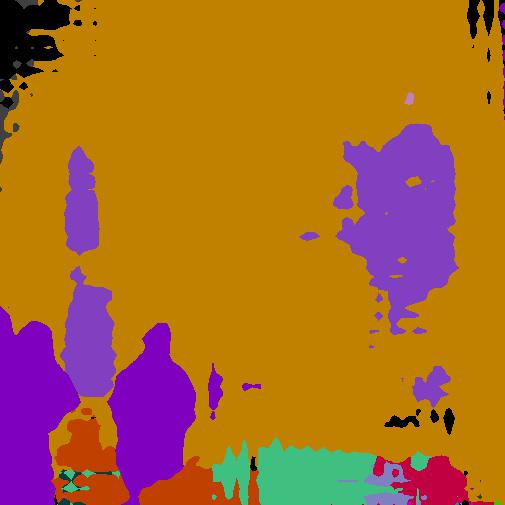} \\

\end{tabular}
\caption{Qualitative results on the PASCAL-Context dataset using 1/8 labeled samples. Failure of our approach. We compare our ('Ours') results with the fully-supervised baseline which is trained only on the labeled subset of data.}
\label{fig:sup_pc3}
\end{figure}

\begin{figure}
\centering

\begin{tabular}{c@{\hspace{1mm}}c@{\hspace{1mm}}c@{\hspace{1mm}}c}
%(a) Original & (b) Ground Truth & (c) Baseline  & (d) Our Results  \\[6pt]

\begin{turn}{90}\hspace{7mm}Orig\end{turn} &
\includegraphics[width=38mm]{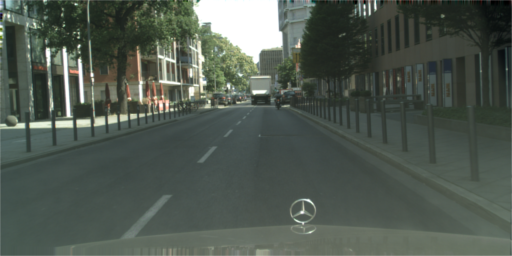} &
\begin{turn}{90}\hspace{7mm}GT\end{turn} &
\includegraphics[width=38mm]{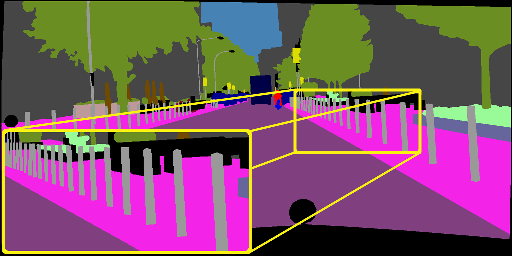} \\
% \vspace{2mm}
\begin{turn}{90}\hspace{7mm}Base\end{turn} &
\includegraphics[width=38mm]{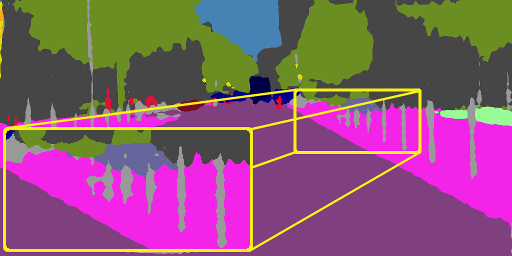} &
\begin{turn}{90}\hspace{7mm}Ours\end{turn} &
\includegraphics[width=38mm]{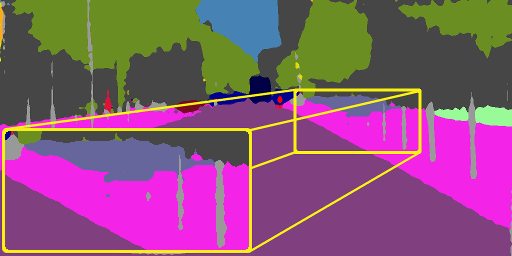} \\

\begin{turn}{90}\hspace{7mm}Orig\end{turn} &
\includegraphics[width=38mm]{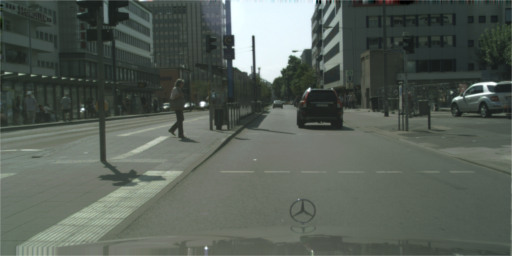} &
\begin{turn}{90}\hspace{7mm}GT\end{turn} &
\includegraphics[width=38mm]{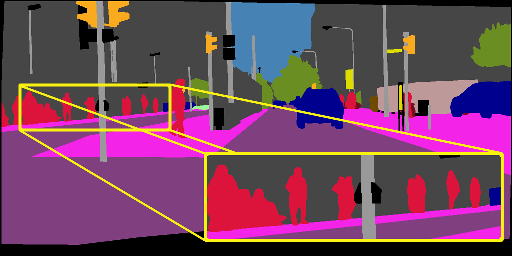} \\
% \vspace{2mm}
\begin{turn}{90}\hspace{7mm}Base\end{turn} &
\includegraphics[width=38mm]{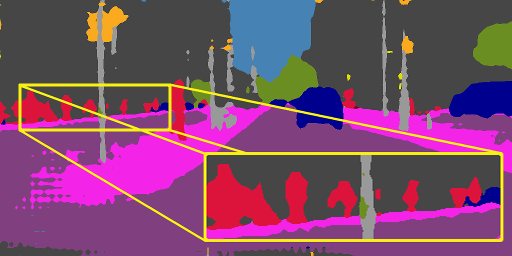} &
\begin{turn}{90}\hspace{7mm}Ours\end{turn} &
\includegraphics[width=38mm]{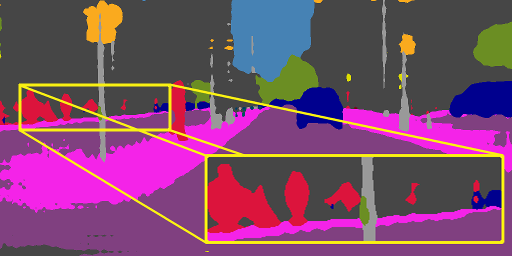} \\

\end{tabular}
\caption{Qualitative results on the Cityscapes dataset using 1/8 labeled samples. Failures of our approach. We compare our ('Ours') results with the fully-supervised baseline (`Base') which is trained only on the labeled subset of data.}
\label{fig:sup_city3}
\end{figure}

In certain situations our method produces imprecise predictions. Sometimes object classes with multiple protrusions like plant leaves, chair legs, etc. are under-segmented by the \igan{} branch, as shown in Figure \ref{fig:fail}(top). Occasionally, our approach can identify certain ambiguous foreground objects as one of the classes, as shown in Figure \ref{fig:fail}(bottom). Also, there exist few cases where some true positive results are wrongly predicted by the classifier. 
However, both qualitative and quantitative results confirm that these failure cases are outweighed by the positive effect of the proposed techniques. In Fig.~\ref{fig:sup_pc3}, we include a few failure cases for PASCAL-context dataset using our approach. Fig.\ref{fig:sup_city3} shows a few failure cases for Cityscapes dataset where few thin objects were not segmented properly using our approach. 

\begin{threeparttable}[htb]
\caption{Ablation study of the contribution of each branch. Results are shown for the 5:95 data split on the PASCAL VOC dataset.}
    \small
    \setlength\tabcolsep{0pt}
\begin{tabular*}{\linewidth}{@{\extracolsep{\fill}} l ccc @{}}
    \toprule
Method          &  \multicolumn{2}{c}{Data}  
                                                            & mIoU\\
        \cmidrule{2-3}                                
                        &labeled(\%)    &unlabeled(\%)\\
        \midrule
DeepLabv2               & 5             & None                & 56.8 \\
\igan{} only            & 5             & 95                  & 60.9 \\
\mt{} only              & 5             & 95                  & 59.0 \\
        \midrule
%\igan{} + \mt{}         & 5             & 95                  & \textbf{62.9} \\
\igan{} + Threshold            & 5       & 95                  & 61.2 \\
\igan{} + Class-wise Threshold & 5       & 95                  & 61.5 \\
\igan{} + CNN                       & 5       & 95                  & 62.2 \\
\igan{} + \mt                       & 5       & 95                  & \textbf{62.9} \\
        \bottomrule
\end{tabular*}

\label{table:ab_branch}
\end{threeparttable}\\

\mypara{ Different \igan{} branch loss terms.}
We trained the generator network with a combination of the cross-entropy (CE) loss, the feature matching (FM) loss, and the self-training (ST) loss.

To justify this configuration, we compare the system performance when using different loss terms; see Table \ref{table:ab_loss}.
There is a consistent performance increase when adding all the proposed loss terms.
We found it crucial for the system stability to train using the FM loss and not the standard GAN loss.

\begin{threeparttable}[htb]
\caption{Ablation study of different GAN loss terms for the generator on the PASCAL VOC dataset. SGAN refers to the standard GAN loss \cite{NIPS2014_5423}, FM refers to the feature-matching loss and ST refers to the self-training loss.}
    \small
    \setlength\tabcolsep{0pt}
\begin{tabular*}{\linewidth}{@{\extracolsep{\fill}} l ccc @{}}
    \toprule
&   \multicolumn{3}{c}{Labeled Data} \\
        \cmidrule{2-4}
Loss Terms                  &1/50
                                        & 1/20
                                                    &1/8\\
        \midrule
CE only                     & 48.3       & 56.8      & 62.0\\
CE + SGAN \cite{NIPS2014_5423}& 54.0     & 57.1      & 62.5\\
%CE + SGAN + ST              & ---       & ---       & --- \\ 
CE + FM                     & 55.4        & 58.4      & 63.9\\
CE + FM + ST                & \textbf{58.1}       & \textbf{60.9} & \textbf{65.4}\\
        \bottomrule
\end{tabular*}

\label{table:ab_loss}
\end{threeparttable}\\

\begin{figure}
\centering
    \includegraphics[width=\columnwidth]{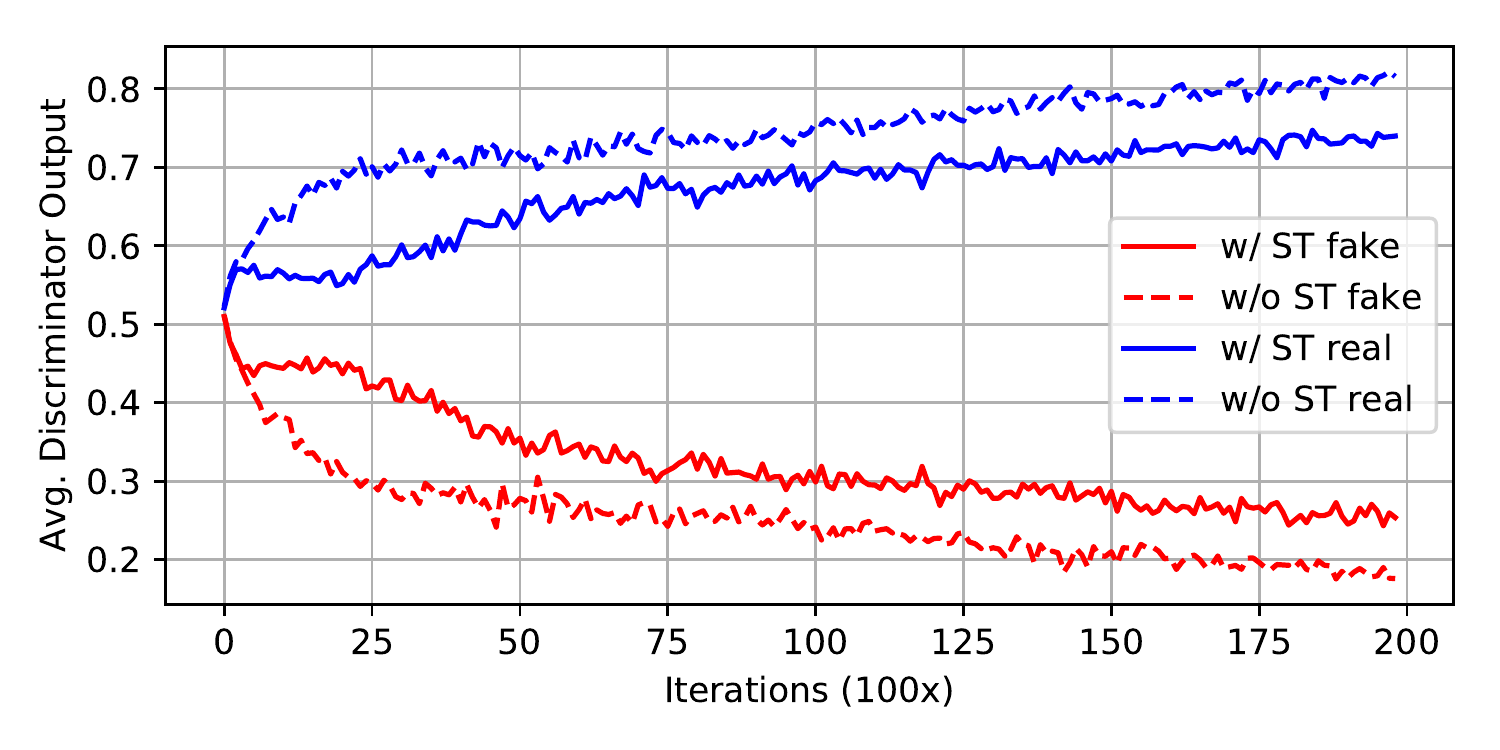}
    \vspace{-5mm}
    \caption{Evolution of the discriminator output during the course of training averaged over real and fake samples separately. Using the self-training loss (w/ ST) prevents $D$ from becoming overly strong and results in better training dynamics compared to the case when self-training is disabled (w/o ST).}
    \label{fig:fm_st}
\end{figure}

Figure~\ref{fig:fm_st} illustrates the effect of using our proposed self-training loss. We plot how the discriminator score changes during the course of training. The scores are averaged over 100 iterations of fake (generated) and real (ground-truth) samples separately. As discussed in Sec.~\ref{sec:training_s}, adding the ST loss impedes the progress of the discriminator and does not allow it to become overly confident, that is, draws its predicted scores towards 0.5. This has a positive effect on the generator performance, in particular with few labeled samples, as can be seen from the last line of Table~\ref{table:ab_loss}.

%%-------------------------------------Study-III------------------------------------------------

\mypara{Semi-supervised multi-label classification.}
In this experiment, we compared the performance of the proposed \mt{} branch with a standard supervised classifier.
Table \ref{table:ab_branch} shows that we already get an improvement of 1.3\% over the \igan{} performance just by using a CNN-based classifier \cite{He_2016_CVPR}, but when we further add the consistency-based semi-supervised classification approach, we observe that the performance improvement increases to 2\%. More detailed comparison between the two classification methods is included in the supplementary file.  
We also conducted a simple heuristic experiment where we deactivate the predicted class channels which have pixel count less than a threshold. In Table \ref{table:ab_branch}, `\igan{} + Threshold' refers to the case where a single threshold is set for all the classes and `\igan{} + Class-wise Threshold' refers to the case where each class has its best respective threshold. We search for the best performing thresholds on the validation set in the range from 1K to 12K pixels at an increment step of 1K.
Figure \ref{fig:ab_voc}(f) and (g) show the effect of adding a CNN-based classifier and an MT-based semi-supervised classifier respectively.

\begin{figure}
\centering
\begin{tabular}{@{}c@{\hspace{1mm}}c@{\hspace{1mm}}c@{\hspace{1mm}}c@{}}
\includegraphics[width=0.5\linewidth]{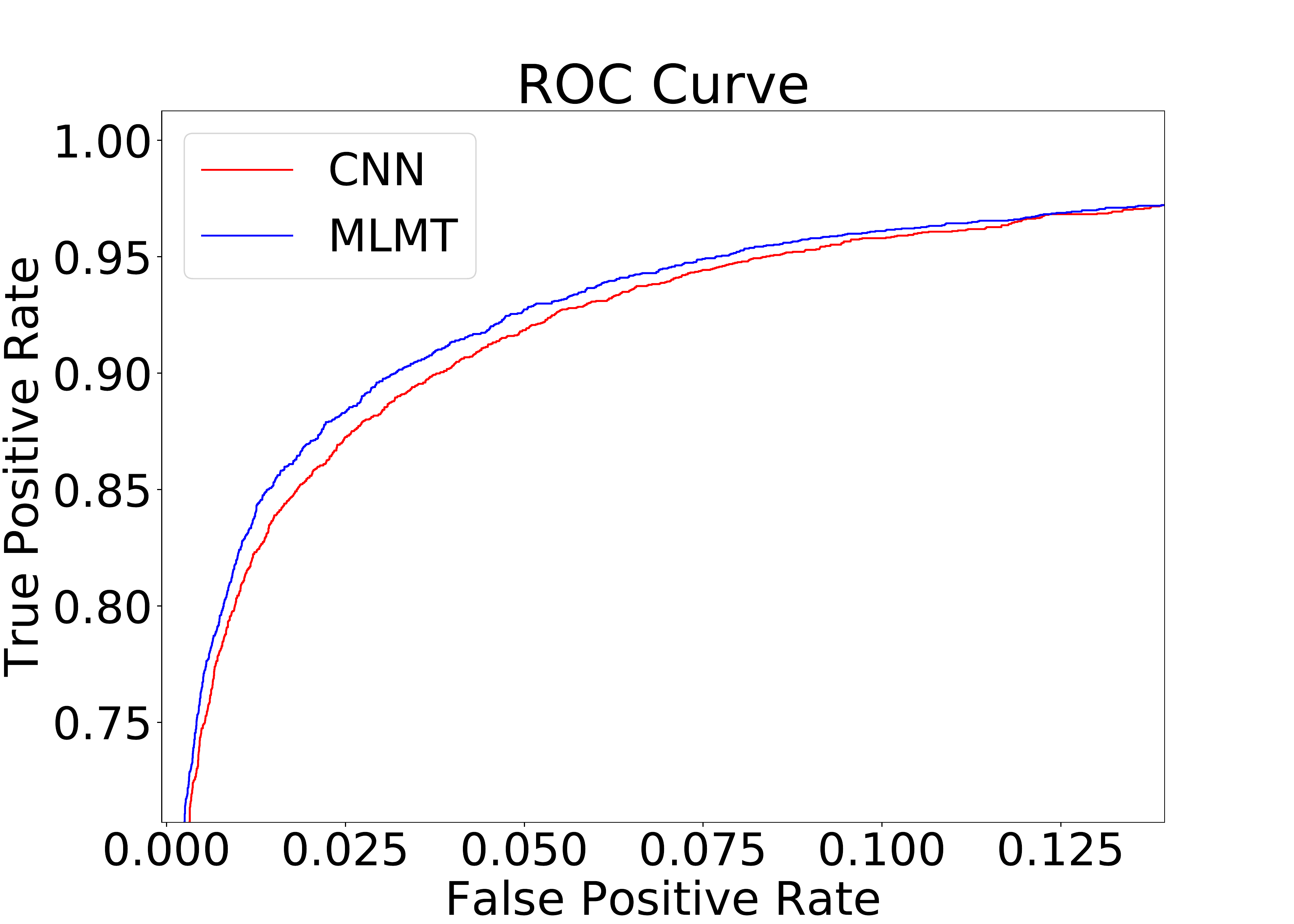} &
\includegraphics[width=0.5\linewidth]{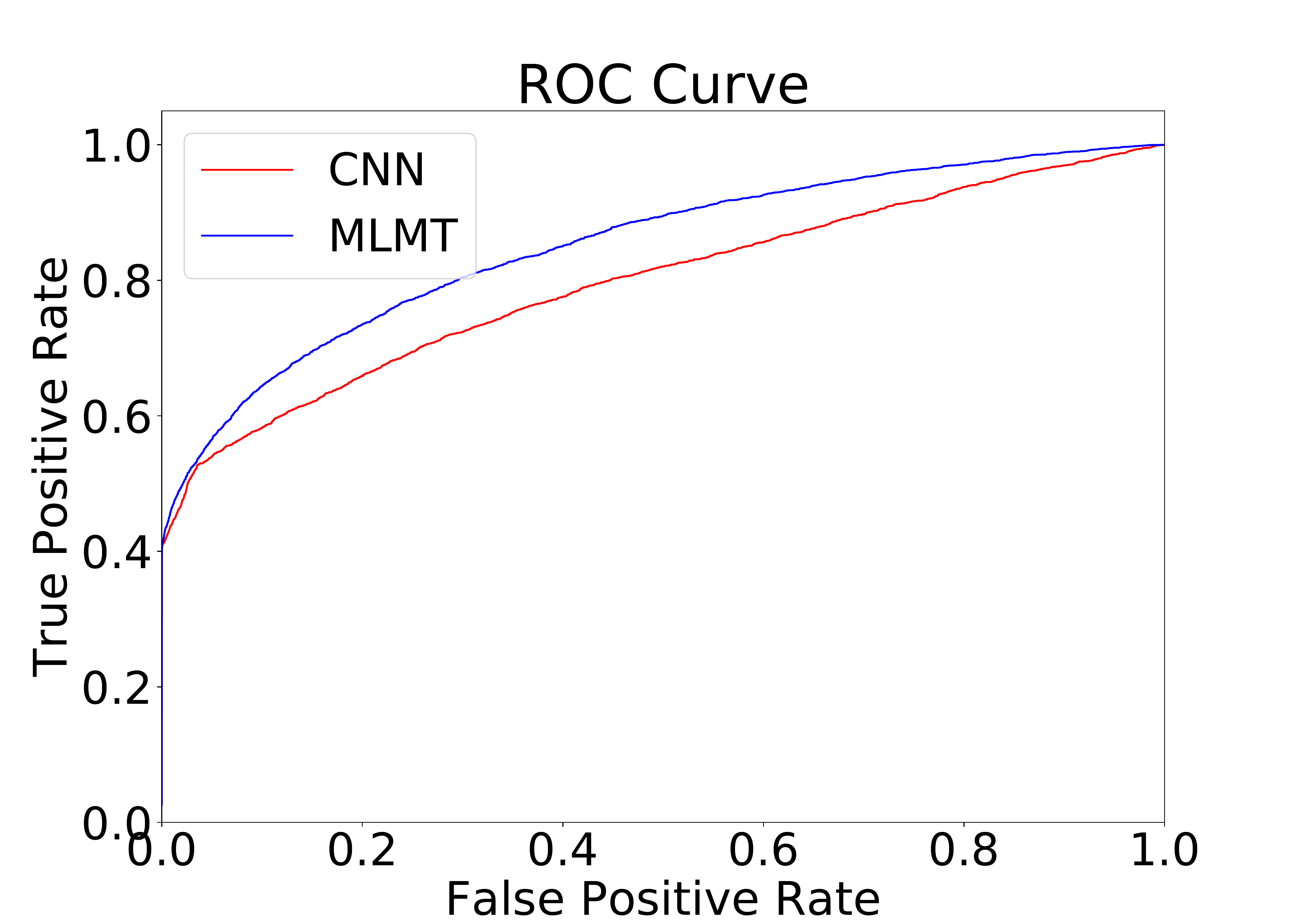} \\
\small (a) & \small (b) 
\end{tabular}
%\vspace{1mm}
\caption{ROC curves for CNN-based classification and MT-based semi-supervised classification method using 5\% labeled data with (a) and without (b) ImageNet pre-training. MT produces fewer false positives, especially when training from scratch.}
\label{fig:roc}

\end{figure}

We also analyze the performance of the CNN-based multi-label classification and MLMT-based semi-supervised multi-label classification independent of the segmentation model. Figure \ref{fig:roc} shows the comparison between the ROC curves of the two methods on the task of multi-label classification. The \mt{} classifier obtains a lower false positive rate for the same true positive rate. The effect is even more pronounced when not using ImageNet pre-training; see Figure \ref{fig:roc}(b). This mode of operation is important for domains where ImageNet pre-training does not help, e.g. bio-medical image analysis.

\begin{threeparttable}[htb]
\caption{Semi-supervised semantic segmentation results on the PASCAL VOC dataset using extra weak image-level annotations. Data splits (\textit{A/B/C}) refers to the usage of \textit{A} pixel-wise labeled samples, \textit{B} image-level labeled samples and \textit{C} unlabeled samples.
% All the invalid cells in the table are marked with (---).
}
    \small
    \setlength\tabcolsep{0pt}
\begin{tabular*}{\linewidth}{@{\extracolsep{\fill}} l ccc @{}}
    \toprule
&   \multicolumn{3}{c}{Data Split (Strong/Weak/Unlab)} \\
        \cmidrule{2-4}
Method                                          & 1.4K/0/9K 
                                                                & 1.4K/9K/0
                                                                                & All/0/0\\ 
        \midrule
DeepLab-CRF-LargeFOV \cite{DBLP:journals/corr/ChenPKMY14}    & 62.5\tnote{c} & ---           & 67.6\tnote{c}\\
WSSL (CRF)\tnote{a} \cite{Papandreou_2015_ICCV} & ---           & 64.6          & ---\\
MDC  \tnote{a} \cite{Wei_2018_CVPR}             & ---           & 62.7          & ---\\
MDC  (CRF)\tnote{a} \cite{Wei_2018_CVPR}        & ---           & 65.7          & ---\\
        \cmidrule{1-4}
DeepLabv2                                       & 65.7          & ---           & 70.7\\
Ours (\igan{} only)\tnote{b}                    & 67.5          & ---           & 71.2\\
Ours (\igan{} + \mt)\tnote{b}                   & 69.6          & \textbf{70.9} & 72.9\\
        \bottomrule
\end{tabular*}

\label{table:wssl}

\begin{tablenotes}[para,flushleft,small]
\item[a] Base network: DeepLab-LargeFOV, \item[b] Base network: DeepLabv2,
\item[c] As reported in \cite{Papandreou_2015_ICCV}
\end{tablenotes}
\end{threeparttable}

%%--------------------------Semi-sup with Weak labels------------------------------

\subsubsection{Semi-supervised Semantic Segmentation with Weak-labels}
To further validate the effectiveness of our approach, we compare it to other semi-supervised segmentation methods \cite{Papandreou_2015_ICCV, Wei_2018_CVPR} that utilize extra weak image-level annotations.
Here, we compare the performance of our approach with methods that use extra image-level annotations \ie 1,464 strongly (w/ segmentation masks) annotated images from the original PASCAL VOC dataset and 9,118 weakly (image-level) annotated images from the augmented SBD dataset. To use extra image-level annotations, we train the \mt{} branch using extra image-level labels for improved multi-label classification. The training procedure and hyperparameters remain exactly same as in the previous semi-supervised setting.  
Table \ref{table:wssl} summarizes the semi-supervised semantic segmentation results with extra $\sim$9K image-level annotations. We achieve an improvement of 5.2\% over the baseline.
Unlike previous methods, our approach does not utilize the CRF post-processing.

%% file: 5_conclusions.tex
\section{Conclusion}

In this work we presented a two-branch approach to the task of semi-supervised semantic segmentation.
The branches are designed to alleviate both low-level and high-level artifacts which often occur when working in a low-data regime.
The effectiveness of this design is demonstrated in a series of extensive experiments.
%An important future work direction is combining the two parts of the current decoupled approach into an end-to-end trainable framework with shared parameters.